\documentclass[10pt,journal,compsoc]{IEEEtran}
\usepackage{booktabs} 
\usepackage{graphicx}
\usepackage{amsmath}
\usepackage{amssymb}
\usepackage{subfigure}
\usepackage{color}
\usepackage{multirow}
\usepackage{threeparttable}
\usepackage{enumitem}
\usepackage{url}

\ifCLASSOPTIONcompsoc
  \usepackage[nocompress]{cite}
\else
  \usepackage{cite}
\fi

\newcommand{\revise}[1]{{#1}}

\begin{document}
%
\title{PaMIR: Parametric Model-Conditioned Implicit Representation for Image-based Human Reconstruction}

\author{Zerong~Zheng,~
        Tao~Yu,~
        Yebin~Liu,~\IEEEmembership{Member,~IEEE} and~Qionghai~Dai,~\IEEEmembership{Senior Member,~IEEE}
\IEEEcompsocitemizethanks{ 
\IEEEcompsocthanksitem Zerong Zheng, Tao Yu, Yebin Liu and Qionghai Dai are with Institute for Brain and Cognitive Sciences, Tsinghua University, Beijing 100084, P. R. China. 
\IEEEcompsocthanksitem Corresponding authors: Yebin Liu and Qionghai Dai. 
}
\thanks{Manuscript received January 31st, 2020.}
}


\markboth{Journal of \LaTeX\ Class Files,~Vol.~14, No.~8, August~2015}%
{Shell \MakeLowercase{\textit{et al.}}: Bare Demo of IEEEtran.cls for Computer Society Journals}

\IEEEtitleabstractindextext{%
\begin{abstract}
Modeling 3D humans accurately and robustly from a single image is very challenging, and the key for such an ill-posed problem is the 3D representation of the human models. To overcome the limitations of regular 3D representations, we propose \textbf{Pa}rametric \textbf{M}odel-Conditioned \textbf{I}mplicit \textbf{R}epresentation (PaMIR), which combines the parametric body model with the free-form deep implicit function. 
In our PaMIR-based reconstruction framework, a novel deep neural network is proposed to regularize the free-form deep implicit function using the semantic features of the parametric model, which improves the generalization ability under the scenarios of challenging poses and various clothing topologies. Moreover, a novel depth-ambiguity-aware training loss is further integrated to resolve depth ambiguities and enable successful surface detail reconstruction with imperfect body reference.
Finally, we propose  a body reference optimization method to improve the parametric model estimation accuracy and to enhance the consistency between the parametric model and the implicit function. With the PaMIR representation, our framework can be easily extended to multi-image input scenarios without the need of multi-camera calibration and pose synchronization. Experimental results demonstrate that our method achieves state-of-the-art performance for image-based 3D human reconstruction in the cases of challenging poses and clothing types. 
\end{abstract}


}

\maketitle

\IEEEdisplaynontitleabstractindextext
\IEEEpeerreviewmaketitle


\IEEEraisesectionheading{\section{Introduction}\label{sec:intro}}
\IEEEPARstart{I}{mage-based} parsing of human bodies is a popular topic in computer vision and computer graphics.
Among all the tasks of image-based human parsing, recovering 3D humans from a single RGB image attracts more and more interests given its wide applications in VR/AR content creation, image and video editing, telepresence and virtual dressing. 
However, as an ill-posed problem, recovering 3D humans from a single RGB image is very challenging due to the lack of depth information and the variations of body shapes, poses, clothing types and lighting conditions. 

\begin{figure}
  \centering
  \includegraphics[width=1.0\linewidth]{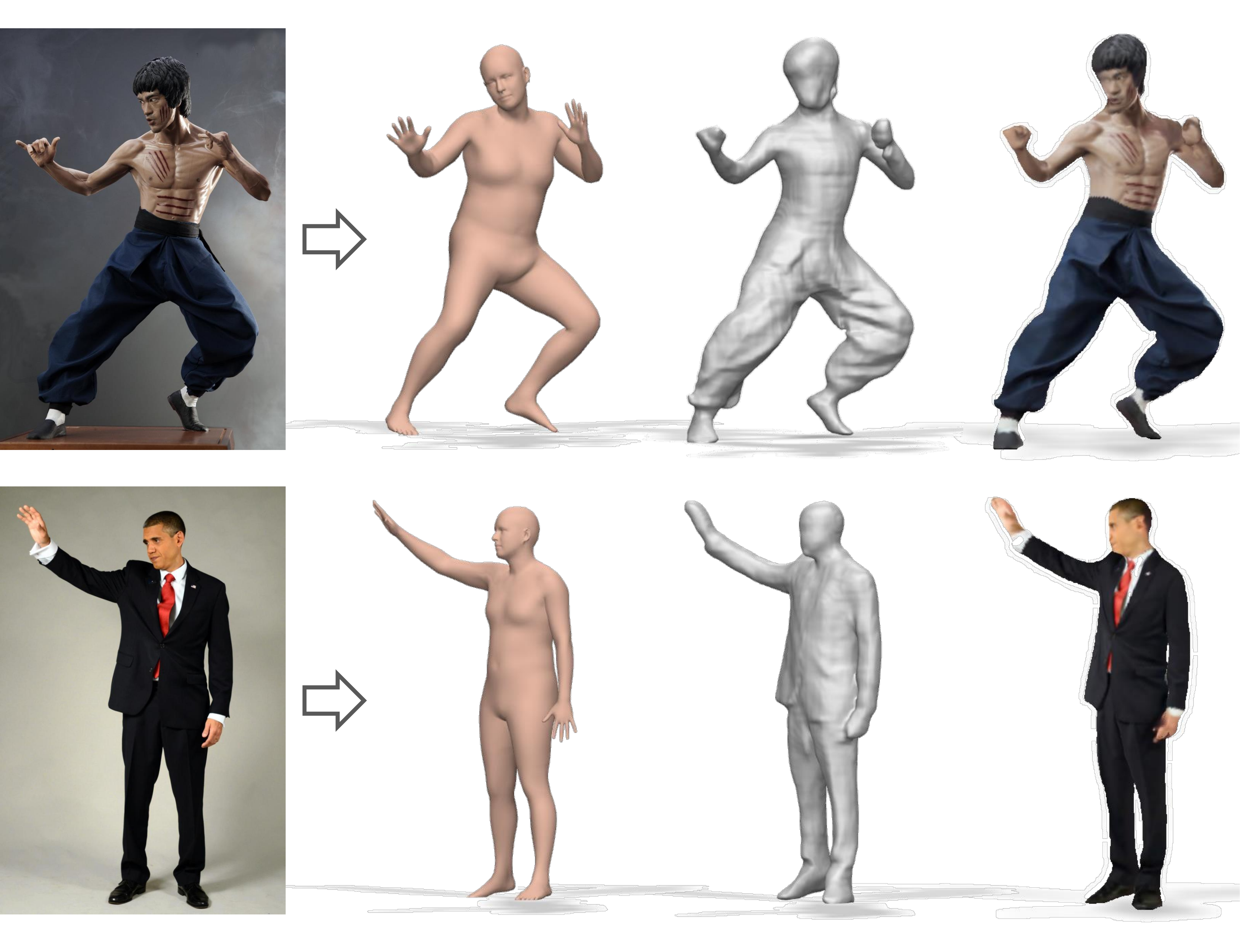}
  \caption{Example results reconstructed by our method. Our method is able to infer the underlying body shape, the 3D surface geometry and its texture given a single RGB image. }
  \label{fig:teaser} 
\end{figure}

\begin{figure*}[!t]
	\begin{center}
		\includegraphics[width=1.0\linewidth]{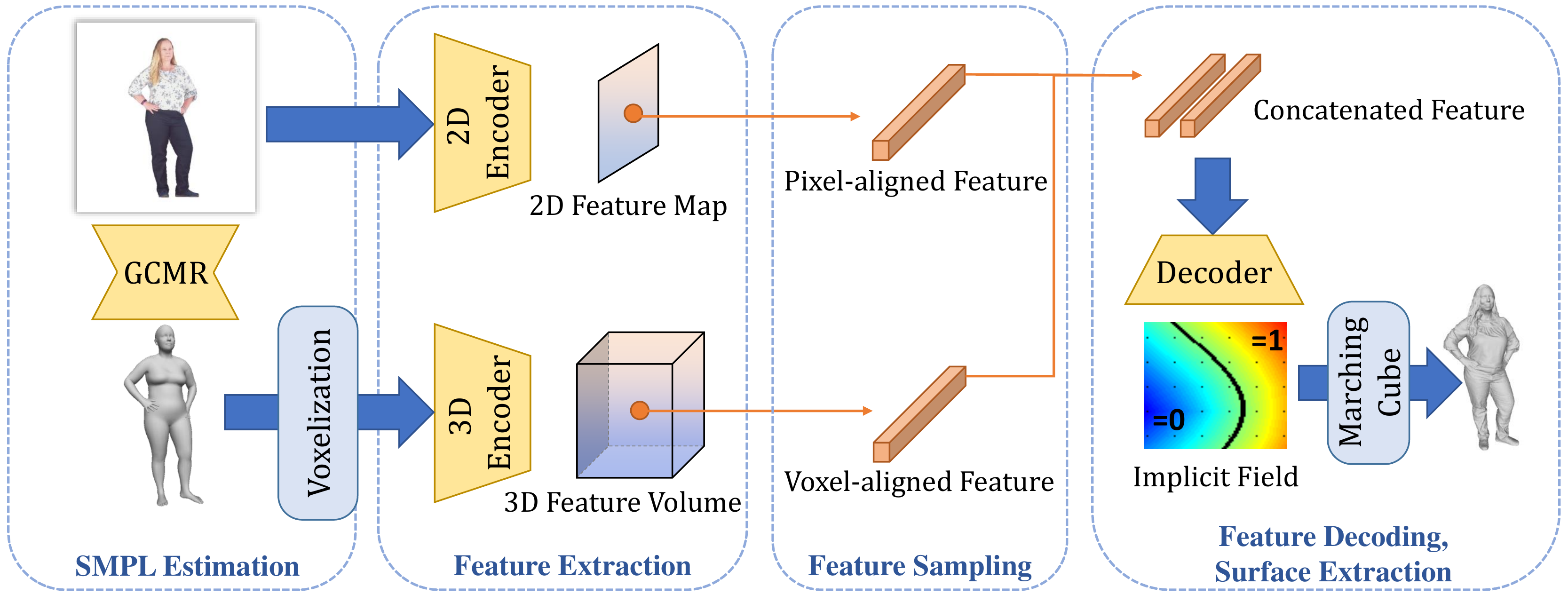}
	\end{center}
	\caption{\revise{Overview of the PaMIR representation and the corresponding network architecture. Given an input image, we first estimate a corresponding SMPL model in the SMPL estimation step. In the following step, the image and the SMPL model are converted into a feature map and a feature volume respectively. Conditioned on the feature vectors sampled in the feature sampling step, the implicit function value for each 3D point can be obtained in the final step.  } }
	\label{fig:arch}
\end{figure*}

Benefiting from the huge progress of deep learning techniques, recent studies have tried to address these challenges using learning-based methods\cite{BodyNet,SiCloPe2019,HMD2019,tex2shape2019,Zheng2019DeepHuman,MouldingHumans2019,FACSIMILE2019,pifuSHNMKL19}. According to their 3D representations, these methods can be roughly classified into two categories: parametric methods and non-parametric methods. Parametric methods like HMD\cite{HMD2019} and Tex2Shape\cite{tex2shape2019} utilize a statistical body template model (e.g., SMPL\cite{SMPL:2015}) as the geometrical prior, and learn to deform the template according to the silhouette and the shading information. 
However, the low dimensional parametric model limits the performance when handling different clothing types such as long dresses and skirts. 
Non-parametric methods use various free-form representations, including voxel grids\cite{BodyNet}, silhouettes\cite{SiCloPe2019}, implicit fields\cite{pifuSHNMKL19} or depth maps\cite{MouldingHumans2019,FACSIMILE2019}, to represent 3D human models. 
Although being able to describe arbitrary clothes, current non-parametric methods still suffer from challenging poses and self-occlusions due to the lack of semantic information and depth ambiguities. 

Recently, DeepHuman\cite{Zheng2019DeepHuman} pioneered in conditioning the non-parametric volumetric representation on the parametric SMPL model, and demonstrates robust reconstruction for challenging poses and diverse garments. 
However, it struggles to recover high-quality geometric details due to the resolution limitation of the regular occupancy volume. More importantly, conditioning the non-parametric representation on SMPL models raises a crucial problem: 
how can we obtain accurate SMPL model estimation given a single image at testing time? 
In DeepHuman, the SMPL models at testing time are estimated using learning-based methods and/or optimization-based methods. Although the recent years have witness a renaissance in the context of single-image body shape estimation\cite{Bogo:ECCV:2016,omran2018NBF,HMR,kolotouros2019cmr,liang2019shapeaware,xu2019denserac}, we observe that there is still an \emph{accuracy gap} between the training and testing stages: during training, we can obtain well-aligned SMPL models for the training data by fitting the SMPL template to the high quality scans, while for testing, the SMPL estimations provided by the state-of-the-art single-image methods\cite{HMR,kolotouros2019cmr} still cannot align with the keypoints or the silhouettes as perfectly as the SMPL models in the training dataset. 


As a result, the neural network trained with the ground-truth image-SMPL pairs cannot generalize well to the testing images which have no ground-truth SMPL annotations. A simple solution is to replace the ground-truth SMPL models with the predicted ones while still using the ground-truth surface scans for training supervision, preventing the network from heavy dependence on accurate SMPL annotations. 
However, due to depth ambiguities, it is impossible to guarantee good alignments between the predicted SMPL models and the ground-truth scans along the depth axis. 
Consequently, by doing so we are forcing the networks to make ``correct'' geometry inference given a ``wrong'' SMPL reference along the depth axis, which will generate even worse results.




To achieve high-quality geometric detail reconstruction while maintaining robustness to challenging pose and clothing styles, in this paper, we propose \emph{Parametric Model-Conditioned Implicit Representation}, dubbed PaMIR, to incorporate the parametric SMPL model and the free-form implicit surface function into a unified learning and optimization framework (Fig.\ref{fig:arch}). The implicit surface function overcomes the resolution limit of volumetric representation, and enables detailed surface reconstruction capability. 
To fill the accuracy gap between the ground truth training data and the inaccurate testing input, we further propose a new depth-ambiguity-aware training loss and a body reference optimization module in the PaMIR-based framework, and achieve surface detail reconstruction even under imperfect body reference initialization.  Specifically, the three technical contributions are summarized as follow:


\begin{itemize}
    
    
\item \textbf{PaMIR.} 
    Our PaMIR representation has the ability to condition the implicit field on the SMPL prediction, which is realized by a novel network architecture converting an image feature map and the corresponding SMPL feature volume into an implicit surface representation. The SMPL feature volume is directly encoded from the SMPL prediction provided by GraphCMR\cite{kolotouros2019cmr}, and serves as a soft constraint for handling challenging poses and/or self-occlusions.

\item \textbf{Training Losses.} 
    We use the predicted SMPL models at both the training and the testing stages. In order to make the network to be more robust against the inaccurate body reference along the depth direction, instead of using a traditional reconstruction loss, we propose a \emph{depth-ambiguity-aware reconstruction loss} that adaptively adjusts the reconstruction supervision based on the current SMPL estimation while respecting 2D image observations. 
    
    
\item \textbf{Body Reference Optimization.} 
    We also propose a \emph{body reference optimization} method for testing process to further refine the body reference. Our optimization method directly utilizes the network itself as well as its outputs to construct an efficient energy function for body fitting. The optimization further alleviates the error caused by the inaccurate initial SMPL estimation, and consequently closing the accuracy gap of SMPL annotation between training and testing. 
\end{itemize}


Overall, in our framework, the underlying body template and the outer surface are mutually beneficial: on one hand, benefiting from the proposed depth-ambiguity-aware reconstruction loss, even the imperfect body template can be used to provide strong semantic information for implicit surface reconstruction; 
on the other hand, the deep implicit function of the outer surface is also used to optimize the underlying body template by minimizing the body fitting error directly. 

Furthermore, thanks to the usage of the common underlying SMPL model as body reference, our PaMIR representation implicitly builds a correspondence relationship across different models. As a result, our method can be easily extended to multi-image setups like multi-view inputs or video inputs without the requirement for  calibration and synchronization. 
Experiments show that our method is able to reconstruct high-quality human models with various body shapes, poses and clothes, and outperforms state-of-the-art methods in terms of accuracy, robustness and generalization capability.

\section{Related Work}
\textbf{Human Reconstruction from multi-view images. }
Previous studies focused on using multi-view images for human model reconstruction \cite{Kanade97,StarckCGA07,LiuTVCG2010}. Shape cues like silhouette, stereo and shading have been integrated to improve the reconstruction performance\cite{StarckCGA07,LiuTVCG2010,WuShadingHuman,WaschbuschWCSG05,VlasicPBDPRM09}. State-of-the-art multi-view real-time systems include  \cite{Fusion4D,Motion2Fusion}. Extremely high-quality reconstruction results have also been demonstrated with tens or even hundreds of cameras \cite{collet2015high}. To capture detailed motions of multiple interacting characters, more than six hundred cameras have been used to overcome the self-occlusion challenges\cite{joo2015panoptic,TotalCapture}.

In order to reduce the difficulty of system setup, human model reconstruction from sparse camera views has recently been investigated by using CNNs for learning silhouette cues \cite{MinimalCam18} and stereo cues \cite{SparseViewHaoLi18}. These systems require about 4 camera views for a coarse-level surface detail capture. Note also that although temporal deformation systems using lightweight camera setups \cite{Vlasic08,Aguiar08,GallSkeleton09} have been developed for dynamic human model reconstruction using skeleton tracking (\cite{Vlasic08,LiuPAMI13}) or human mesh template deformation \cite{Aguiar08}, these systems require a pre-scanned subject-specific 3D template for deformation optimization.   \newline

\noindent\textbf{Image-based Parametric Body Estimation. }
Dense 3D parsing from a single image has attracted substantial interest recently because of the emergence of human statistical models like SCAPE \cite{SCAPE} and SMPL\cite{SMPL:2015}. For example, by fitting the SMPL model to the 2D keypoint detections\cite{OpenPoseCao} and other dense shape cues\cite{DensePose}, the shape and pose parameters can be automatically obtained from a single image\cite{Bogo:ECCV:2016,UnitethePeople2017}. Instead of optimizing mesh and skeleton parameters, recent approaches proposed to train deep neural networks that directly regress the 3D shape and pose parameters from a single image \cite{omran2018NBF,HMR,kolotouros2019cmr,liang2019shapeaware,xu2019denserac}. 
\revise{The estimation accuracy of these methods are further improved by performing fitting optimization after network inference\cite{HoloPose2019}, introducing model optimization into the training loop\cite{SPIN2019}, incorporating adversarial prior in temporal domain\cite{VIBE2020}, or combining global and local features to produce fine-grained human body poses\cite{Pose2Pose2020}. }
However, the parametric body models can only capture the shape and pose of a minimally clothed body, thus lack the ability to represent a 3D human model with more general clothing layers.
\newline


\noindent\textbf{Non-parametric Human Reconstruction from a Single Image.}
Regarding single-image human model reconstruction using non-parametric models, recent studies have adopted techniques based on silhouette estimation\cite{SiCloPe2019}, template-based deformation\cite{HMD2019,tex2shape2019}, depth estimation\cite{MouldingHumans2019,FACSIMILE2019} and volumetric reconstruction\cite{BodyNet,Zheng2019DeepHuman} . Although they have achieved promising results, typical limitations still exist when using a single 3D representation: silhouette-based methods like \cite{SiCloPe2019} is suffer from lack of details and view inconsistency,  template-based deformation methods\cite{HMD2019,tex2shape2019} are unable to handle loose clothes, 
depth-based methods\cite{MouldingHumans2019,FACSIMILE2019} cannot handle self-occlusions naturally, 
and volumetric methods\cite{BodyNet,Zheng2019DeepHuman} cannot recover high-frequency details due to their cubically growing memory consumption. 
\newline



\noindent\textbf{Implicit Representation.}
How to represent 3D surface is a core problem in 3D learning. Explicit representations like point clouds\cite{qi2016pointnet,lin2018learning}, voxel grids\cite{maturana2015voxnet,su2015multi}, triangular meshes\cite{groueix2018atlasnet,wang2018pixel2mesh} have recently been explored to replicate the success of deep learning techniques on 2D images. However, the loss of structure information in point cloud, the memory requirement of regular voxel grid, and the fixed topology of meshes make these explicit representations unsuitable to represent arbitrary high-quality 3D surfaces in neural networks.
Implicit surface representations\cite{carr2001reconstruction,curless1996volumetric}, on the other hand, overcome these challenges and demonstrate the best flexibility and expressiveness for topology representation. It defines a surface as the level set of a function of occupancy probability or surface distance. Recent works \cite{park2019deepsdf,chen2019LearningImplicitFeild,Mescheder2019OccupancyNetwork,chibane20ifnet,local_implicit} have shown promising results on generative models for 3D object shape inference based on deep implicit representations. \newline

\noindent\textbf{Deep Implicit Representation for Human Reconstruction.}
\revise{The success of deep implicit representations in general object modeling has inspired research in 3D human reconstruction\cite{pifuSHNMKL19,pifuhd2020,ARCH2020,Monoport2020,bhatnagar2020ipnet}. 
For example, PIFu~\cite{pifuSHNMKL19} proposed to regress an deep implicit function using pixel-aligned image features and is able to reconstruct high-resolution results, which is then accelerated to real-time framerate in \cite{Monoport2020}. PIFuHD~\cite{pifuhd2020} extended PIFu to capture more local details by adding a fine-level feature extraction network. 
However, both PIFu and PIFuHD are prone to reconstruction artifacts in cases of challenging poses and self-occlusions. 
In contrast, our method achieve more robust performance under these challenging scenarios by introducing SMPL model as an additional semantic condition for the implicit representation. 
ARCH\cite{ARCH2020} and IP-Net\cite{bhatnagar2020ipnet} are two concurrent works that also use SMPL model for semantic reconstruction. In particular, ARCH\cite{ARCH2020} proposed to regress animation-ready 3D avatars in a canonical pose (A-pose), but fails to generate accurate results, especially for loose clothes and human-object interaction (e.g., holding a camera with two hands as in Fig.\ref{fig:results}) because it is ambiguous to determine the position of the objects and accessories in A-pose.    
IP-Net\cite{bhatnagar2020ipnet} jointly predict the outer 3D surface of a human model, the inner body surface and the body part labels, but is restricted to point cloud inputs. 
In contrast, our method can naturally handle human-object interaction and can be flexibly adopted in both single-image and multi-image setups. }




\section{Surface Representation}
\label{sec:surface_representation}
\subsection{Deep Implicit Function}
\label{sec:surface_representation:implicit_fields}
A deep implicit function defines a surface as the level set of an occupancy probability function $F$, e.g. $F(p) = 0.5$, where $p\in\mathbb{R}^3$ denotes a 3D point and $F: \mathbb{R}^3 \mapsto [0, 1]$ is represented by a deep neural network. In practice, in order to represent the surface of a specific object using a general neural network, the function $F$ also takes a condition variable (e.g., image feature of the object) as input and thus can be written as: 
\begin{equation}
    F(p, c): \mathbb{R}^3\times \mathcal{X} \mapsto [0, 1]
\end{equation}
where $c\in\mathcal{X}$ is the condition variable that encodes the overall shape of a specific surface and can be custom designed in accordance of applications\cite{chen2019LearningImplicitFeild,park2019deepsdf,Mescheder2019OccupancyNetwork}. Intuitively, $F$ predicts the continuous inside/outside probability field of a 3D model, in which iso-surface can be easily extracted

In PIFu\cite{pifuSHNMKL19}, the authors combined the condition variable with the point coordinate and formulate a pixel-aligned implicit function as:
\begin{equation}
\label{eqn:pifu}
    F(C(p)): \mathbb{R}^3 \mapsto [0, 1]
\end{equation}
\begin{equation}
\label{eqn:pifu2}
    C(p) = \left(S\left(\mathbf{F}_{I}, \pi(p)\right), Z(p)\right)^\mathrm{T}
\end{equation}
where $\mathbf{F}_I = E_I (\mathbf{I})$ represents the image feature map from the deep image encoder $E_I(\cdot)$, $\pi(p)$ the 2D projection of $p$ on the feature map $\mathbf{F}_I$, $S\left(\cdot, \cdot\right)$ is the sampling function used to sample the value of $\mathbf{F}_I$ at pixel $\pi(p)$ using bilinear interpolation, and $Z(p)$ is the depth value of $p$ in the camera coordinate space. A weak perspective camera is assumed in both PIFu and this paper. 
Intuitively, given a pixel on the image, PIFu casts a ray through that pixel along the direction of z-axis and estimates the values of occupancy probability on that ray based on the local image feature of the given pixel. Thanks to the usage of pixel-level features as condition variable, PIFu can reconstruct fine-scale detailed surfaces that are well aligned to the input images. 

PIFu\cite{pifuSHNMKL19} demonstrates high-quality human digitization for fashion poses, like standing or walking. However, we argue that purely relying on 2D image feature is insufficient to handle severe occlusions and large pose variations for single-image 3D human recovery, especially when ideal high quality dataset (covering sufficient poses, shapes and cloth types) is inaccessible. Specifically, the reasons are two-folds. On one hand, under the scenarios of self-occlusions, there may be multiple peaks of occupancy probability along the z-rays, but it is hard to infer these types of changes consistently and accurately only based on the local pixel-level 2D features. 
On the other hand, without the awareness of the underlying body shape and pose, the reconstruction results are prone to seriously incomplete or asymmetric bodies, like breaking arms and unequal legs, in the cases of challenging poses. 

\subsection{Parametric body Model}
\label{sec:surface_representation:smpl}
Inspired by DeepHuman\cite{Zheng2019DeepHuman}, we integrate the parametric body model, SMPL \cite{SMPL:2015}, to regularize the human reconstruction. 
SMPL is a function $M(\cdot)$ that maps pose $\theta$ and shape $\beta$ to a mesh of $n_{S}$ vertices:
\begin{equation}
\begin{gathered}
    M(\beta, \theta) = W(T(\beta, \theta), J(\beta), \mathbf{W})) \\
    T(\beta, \theta) = \mathbf{T} + B_s(\beta) + B_p(\theta)  
\end{gathered}
\end{equation}
where linear blend-skinning $W(\cdot)$ with skinning weights $\mathbf{W}$ poses the T-pose template $\mathbf{T} + B_s(\beta) + B_p(\theta) $ based on its skeleton joints $J(\cdot)$. As shown in DeepHuman, a SMPL model fitted to the input image can serve as a strong geometry prior, and guarantee pose and shape aware human model reconstructions. However, estimating SMPL model given a single image is a fundamentally ill-posed problem. 
Recent studies have explored to learn from large-scale datasets to address this challenge\cite{omran2018NBF,HMR,kolotouros2019cmr,liang2019shapeaware,xu2019denserac,HoloPose2019,SPIN2019,VIBE2020,Pose2Pose2020}. 
\revise{In this work we use GCMR\cite{kolotouros2019cmr}, one of the state-of-the-art networks as the backbone for body shape inference. Note that GCMR can be replaced with other equivalent networks such as SPIN\cite{SPIN2019} for single images and VIBE\cite{VIBE2020} for videos, as we do not make any assumptions on how the initial body is estimated and our method is able to deal with inaccurate body initialization thanks to our body reference optimization (Sec.\ref{sec:method:optm}).}

\begin{figure*}
	\begin{center}
		\includegraphics[width=1.0\linewidth]{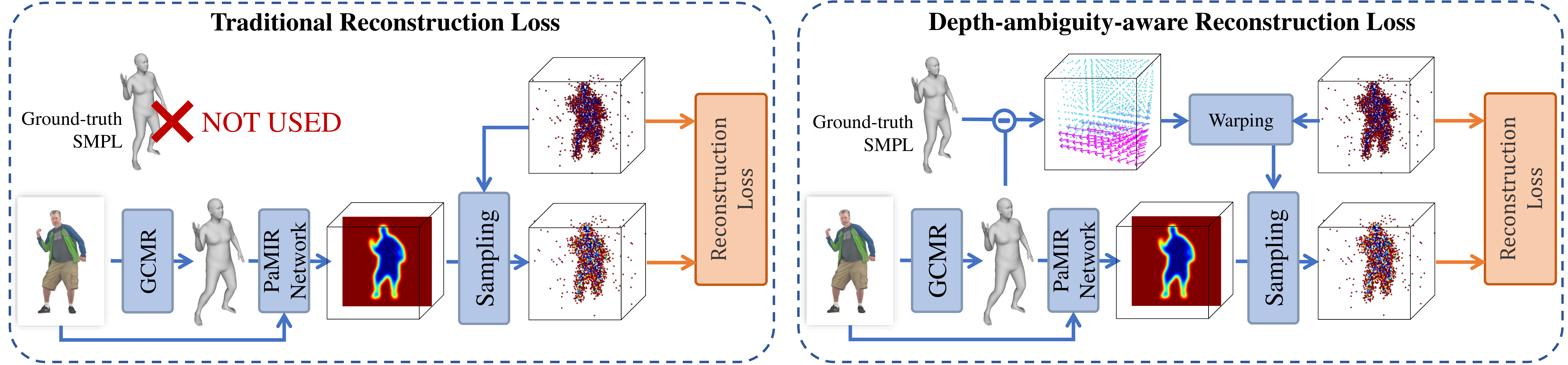}
	\end{center}
	\caption{Comparison of the traditional reconstruction loss and our proposed depth-ambiguity-aware reconstruction loss. Unlike the traditional reconstruction loss, we shift the point samples according to the depth difference between the ground-truth SMPL and the predicted one. In this way the network is able to learn to infer an implicit field registered with the predicted 3D pose. }
	\label{fig:adaptive_loss}
\end{figure*}

\subsection{PaMIR: Parametric Model-Conditioned Implicit Representation}
\label{sec:surface_representation:pair}
To combine the strengths of parametric body models and non-parametric implicit field, we introduce Parametric Model-Conditioned Implicit Representation (PaMIR). Specifically, in PaMIR, we define the definition of $C(p)$ in Eqn.(\ref{eqn:pifu2}) as: 
\begin{equation}
\begin{gathered}
\label{eqn:pair}
    C(p) = \left(S\left(\mathbf{F}_{I}, \pi(p)\right), S\left(\mathbf{F}_{V}, p\right)\right)^\mathrm{T} 
\end{gathered}
\end{equation}
where $S\left(\mathbf{F}_{I}, \pi(p)\right)$ has the same definition as Eqn.(\ref{eqn:pifu2}), while $\mathbf{F}_{V}$ is the feature volume and $S\left(\mathbf{F}_{V}, p\right)$ represents the voxel-aligned volumetric feature at $p$ sampled from $\mathbf{F}_{V}$. The feature volume $\mathbf{F}_{V}$ is obtained by firstly converting the SMPL mesh into an occupancy volume $\mathbf{V}_O$ through mesh voxelization and then encoding it through a 3D encoder $E_V$, i.e., $\mathbf{F}_{V} = E_V(\mathbf{V}_O)$. Given the per-pixel feature vector $S\left(\mathbf{F}_{I}, \pi(p)\right)$ of $p$ as well as its per-voxel feature vector $S\left(\mathbf{F}_{V}, p\right)$, we learn an implicit function $F(C(p))$ in Eqn.(\ref{eqn:pifu}) that can classify whether $p$ is inside or outside the surface.
Note that although the introduction of 3D dense feature volume may increase memory consumption, in practice, we found that a relatively small feature volume is enough for soft semantic regularization. 
Moreover, we omit $Z(p)$ in $C(p)$ since $S\left(\mathbf{F}_{V}, p\right)$ already contains 3D coordinate information. The illustration of the proposed PaMIR representation and the corresponding network is shown in Fig.\ref{fig:arch}. 

In PaMIR, the free-form implict representation is regularized by the semantic features of the parametric model. 
As we can see in the following sections, introducing SMPL model as a body reference has several advantages: 
\begin{enumerate}
    \item Pose Generalization. For surface geometry reconstruction, the SMPL input can be regarded as an initial guess for the network, which helps to factor out pose changes of the subject and eliminate depth ambiguities, thus making the network mainly focusing on surface detail reconstruction. 
    As a result, our method has more robust performance compared with PIFu especially under challenging poses. 
    \item Multi-modal Prediction. Unlike PIFu\cite{pifuSHNMKL19} or Occupancy Network\cite{Mescheder2019OccupancyNetwork} that only condition on the input images, our PaMIR representation also conditions on the underlying bodies. Thus, in contrast to reconstruct one model for each input image, our method can reconstruct various plausible models given different but plausible body poses (Fig.\ref{fig:pair1}).
    \item Easy Extension to Multi-image Setups. With the common underlying SMPL model as a body reference, our PaMIR representation also implicitly builds a correspondence relationship across different models and images. As a result, our method can be easily extended to multi-image settings like multi-view input and video input without explicit calibration and synchronization.
\end{enumerate}

\section{Method}
\label{sec:method}

\subsection{Overview}
\label{sec:method:network}

Our PaMIR-based 3D human reconstruction is implemented as a neural network. An overview of the network is illustrated in Fig.\ref{fig:arch}. 
Given a single image $\mathbf{I}$ as input, our method first feeds it into the GCMR network to estimate an initial SMPL model, which is then converted into an occupancy volume through voxelization. In the feature extraction step, the input image is encoded into a feature map $\mathbf{F}_{I}$ by a 2D convolution network, while the occupancy volume are encoded into a feature volume $\mathbf{F}_{V}$ by a 3D convolution network. For each point in the 3D space, its pixel-aligned image feature $S\left(\mathbf{F}_{I}, \pi(p)\right)$ and voxel-aligned volume feature $S\left(\mathbf{F}_{V}, p\right)$ are sampled in the feature map and the feature volume,  respectively. The two feature vectors are then concatenated and translated to an occupancy probability value by a feature-to-occupancy decoder as formulated in Eqn.(\ref{eqn:pifu}) and (\ref{eqn:pair}). 

To alleviate the reliance on accurate SMPL annotations, we use the training images, the SMPL models estimated by GCMR and the corresponding ground-truth meshes to train the other parts of the network. To deal with the depth inconsistency between the predicted SMPL models and the ground-truth scans, we carefully design a depth-ambiguity-aware reconstruction loss as elaborated in Sec.\ref{sec:method:loss}. Moreover, for inference, we propose body reference optimization, and optimize the human body template as well as the implicit function in an iterative manner in Sec.\ref{sec:method:optm}. To obtain the final reconstruction results, we densely sample the occupancy probability field over the 3D space and extract the iso-surface of the probability field at threshold 0.5 using the Marching Cube algorithm. Texture inference can be performed in a similar way to geometry inference (Sec.\ref{sec:method:texture}). Our framework can be easily extended to multi-image setups, which is described in Sec.\ref{sec:method:multiimage}.

\subsection{Depth-ambiguity-aware Reconstruction Loss}
\label{sec:method:loss}

To train our network, we sample 3D points in 3D space around the human model, infer their occupancy probabilities and construct a per-point reconstruction loss. The traditional reconstruction loss is defined as the mean square error between the predicted occupancy probability of the point samples and the ground-truth ones. However, we argue that with a single-image setup, depth ambiguity is inevitable for many body poses and hence it is unpractical to force the network output to be perfectly identical with the ground-truth. Furthermore, if we utilize the predicted SMPL models as reference of 3D shape and pose, using traditional reconstruction loss will lead to negative impact on the reconstruction accuracy because the predicted models may not be well aligned with the ground-truth along the z-axis. To deal with this issue, we propose a novel depth-ambiguity-aware reconstruction loss defined as:
\begin{equation}\label{eqn:reconstruction_loss}
    \mathcal{L}_R = \frac{1}{n_p}\sum_{i=1}^{n_p}  \left| F(C(p_i+ \Delta p_i)) - F^{*}(p_i) \right|^2, 
\end{equation}
where $n_p$ is the number of point samples, $p_i$ a 3D point sample indexed by $i$, $F^{*}(p_i)$ the ground-truth occupancy value of $p_i$, and $\Delta p_i = (0, 0, \Delta z_i )^\mathrm{T}$ is the compensating translation along z-axis. $\Delta z_i$ is calculated on-the-fly during network training using the explicit correspondences between the predicted SMPL model and the ground-truth SMPL model according to
\begin{equation}
\label{eqn:z_translation}
    \Delta z_i = \sum_{j\in\mathcal{N}(i)} \frac{\omega_{j\rightarrow i}}{\omega_i} \left(Z(v_j) - Z(v^{*}_j)\right), 
\end{equation}
where $\mathcal{N}(i)$ is the nearest SMPL vertex set of $p_i$ and  $\lvert\mathcal{N}(i)\rvert = 4$, $\omega_{j\rightarrow i}$ the corresponding blending weight, $\omega_i$ the weight normalizer, $v_j$ and $v^{*}_j$ are the $j$-th vertex of the predicted SMPL model and the corresponding ground-truth SMPL model, respectively. The blending weight is defined according to the distance between the point sample $p_i$ and its neighboring SMPL vertex $v_j$: 
\begin{equation}
\label{eqn:blending_weights}
\begin{gathered}
    \omega_{j\rightarrow i} = \exp\left({-\frac{\lVert p_i -v_j  \rVert}{2\sigma^2}}\right), \\
    \omega_i = \sum_{j\in\mathcal{N}(i)} \omega_{j\rightarrow i}. 
\end{gathered}
\end{equation}
Intuitively, with the depth-ambiguity-aware loss, we are guiding the network to output plausible surface corresponding to the predicted SMPL model but not the exact ground-truth occupancy volume. In this way the network is able to learn to infer an implicit field registered with the predicted 3D pose. 
The illustration of depth-ambiguity-aware reconstruction loss is shown in Fig.\ref{fig:adaptive_loss}.




\subsection{Body Reference Optimization}
\label{sec:method:optm}
\begin{figure}
	\begin{center}
		\includegraphics[width=1.0\linewidth]{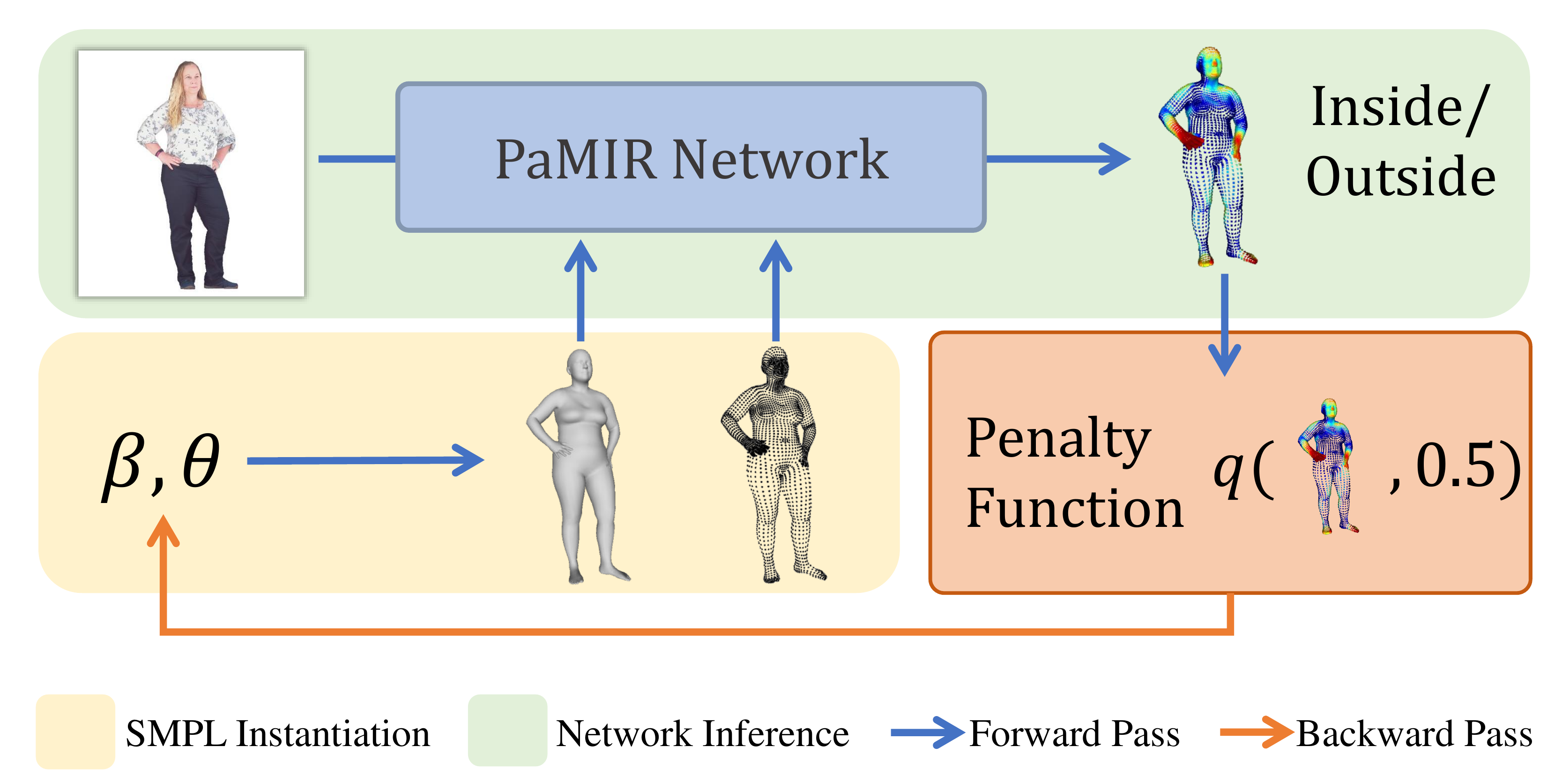}
	\end{center}
	\caption{Illustration of body reference optimization. The estimated SMPL parameters are firstly decoded into a SMPL mesh and then the occupancy values of its vertices are infered by our PaMIR network. Through minimizing the body fitting loss, the SMPL parameters are optimized iteratively. }
	\label{fig:optm}
\end{figure}
Although our training scheme already prevents our network from being heavily dependent on the accuracy of SMPL estimation, the inconsistency between the image observation and the SMPL estimation may still lead to reconstruction artifacts. Fortunately, the predictions of our method can be further refined at inference time. Specifically, we can improve the accuracy of SMPL estimation by minimizing:
\begin{equation}
    (\beta^*, \theta^*) = \arg\min_{\beta, \theta} \mathcal{L}_{B} + \lambda_R \mathcal{L}_{REG}
\end{equation}
where $\mathcal{L}_{B}$ is the body fitting loss used to encourage the alignment of the predicted implicit function and SMPL model and $\mathcal{L}_{REG}$ is a regularization term penalizing the difference between $(\beta^*, \theta^*)$ and the initial prediction. The body fitting loss is defined as following:
\begin{equation}
\label{loss:body_fitting}
    \mathcal{L}_{B} = \frac{1}{n_{S}}\sum_{j=1}^{n_{S}}  q\left(F\left(c(v_j)\right)- 0.5\right)
\end{equation}
where $n_{S}$ is the number of SMPL vertices,  $v_j$ is the $j$-th predicted SMPL vertex and $q(\cdot)$ is the penalty loss defined as:
\begin{equation}
    q(x) = 
    \begin{cases}
    |x| & x \geq 0, \\
    \frac{1}{\eta} |x| & x < 0, 
    \end{cases}
\end{equation}
where $\eta = 5$. 
In other words, $p(\cdot)$ applies more penalization on the vertices that fall outside of the predicted surface while allowing the body to shrink into the surface in case of loose clothes like skirts and dresses. The regularization term is defined as
\begin{equation}
    \mathcal{L}_{REG} = \lvert\beta-\beta_{init}\rvert_2^2 + \lvert\theta-\theta_{init}\rvert_2^2, 
\end{equation}
where $(\beta_{init}, \theta_{init})$ is the initial SMPL parameters estimated by GCMR. Note that other constraints like 2D keypoint detection results can also be added to further improve the fitting performance. 

The insight behind the formulation of body fitting loss  is that the predicted SMPL may not be perfectly aligned with the image observation and consequently, the output implicit function is a compromise between these two information. Thus, by minimizing the body fitting loss, we can eliminate the inconsistency between the SMPL prediction and the image observation. As a result, more consistent SMPL prediction as well as more accurate surface inference can be obtained. The idea of our body reference optimization is illustrated in Fig.\ref{fig:optm}. 

\begin{figure}
	\begin{center}
		\includegraphics[width=1.0\linewidth]{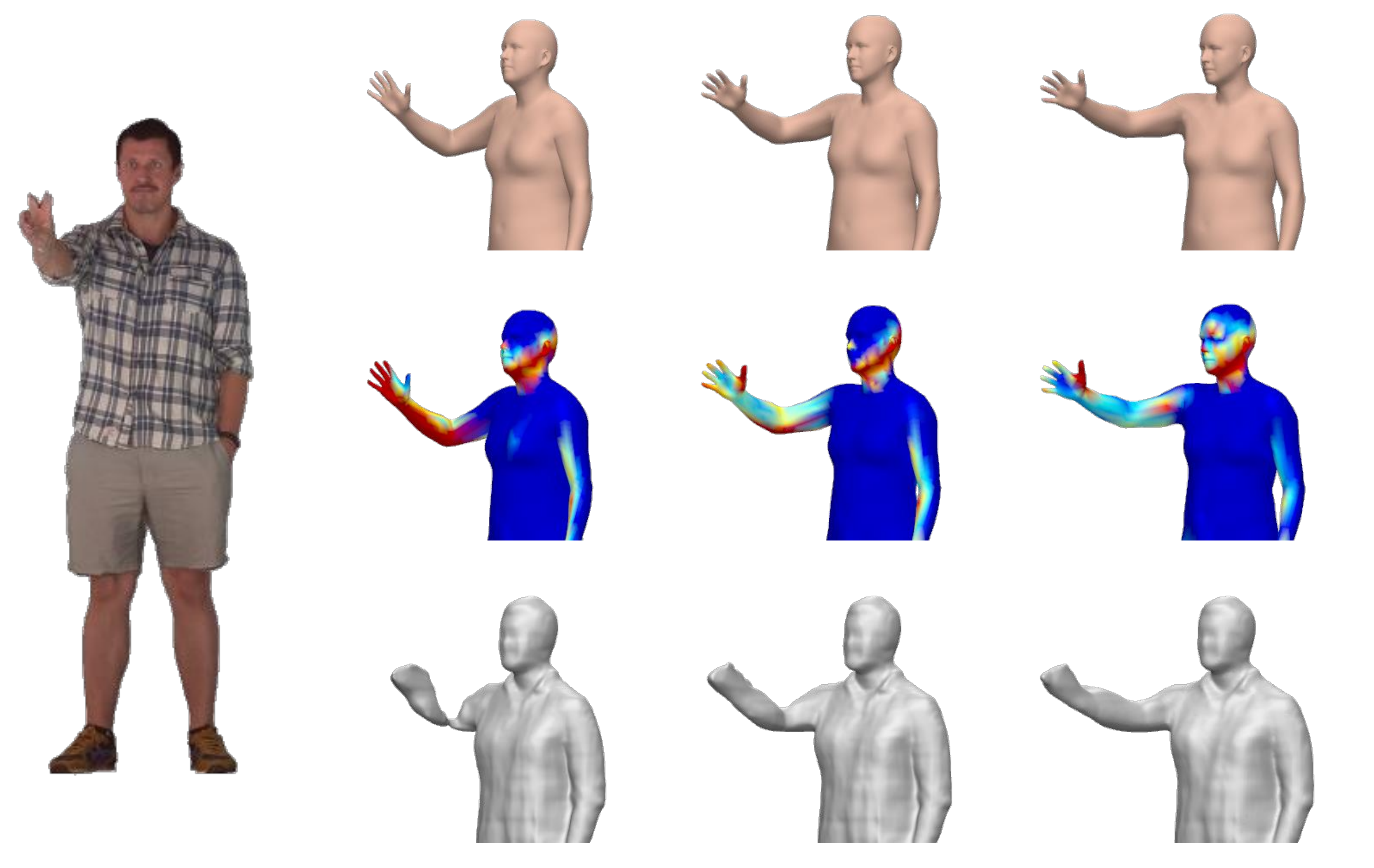}
	\end{center}
	\caption{Visualization of the body optimization process. The leftmost column: the input image. The 2nd to 4th columns: the reconstruction results before reference body optimization, the intermediate status of optimization and the results after optimization. We present the SMPL models in the top row and the reconstructed models in the bottom row. The median row shows the occupancy probability of SMPL vertices: blue color represents $F(C(v)) = 1.0$, while red means  $F(C(v)) = 0.0$.  }
	\label{fig:vis_body_optimization}
\end{figure}

\revise{Although body model optimization has been proposed in SMPLify\cite{Bogo:ECCV:2016} and HoloPose\cite{HoloPose2019}, our optimization scheme are substantially different from that in previous works. Existing methods fit body models to image observation such as 2D keypoints detection, 3D keypoints estimation and/or dense correspondences, while ours directly utilizes the human reconstruction network to fit SMPL model into the implicit surface via minimizing the loss in Eqn.(\ref{loss:body_fitting}). Our scheme guarantees that the image observation, the estimated body model and the reconstructed outer surface are aligned with each other, and eliminates the requirements of sparse/dense keypoint detection (although they can be used as additional constraints). }

We visualize the optimization process in Fig.\ref{fig:vis_body_optimization}. In this figure, the initial pose of the right arm is incorrect, which leads to reconstruction artifacts. However, as the optimization is carried out, the right arm gradually moves towards the right position. In the meantime, the reconstructed mesh becomes more and more plausible. To clarify, the reconstruction results in this experiment do not contradict the results in Fig.\ref{fig:eval_endtoend}: in this experiment, the right arm is almost invisible and consequently it is difficult for the network to infer the correct geometry with an inaccurate arm pose.

\begin{figure}
    \centering
    \includegraphics[width=\linewidth]{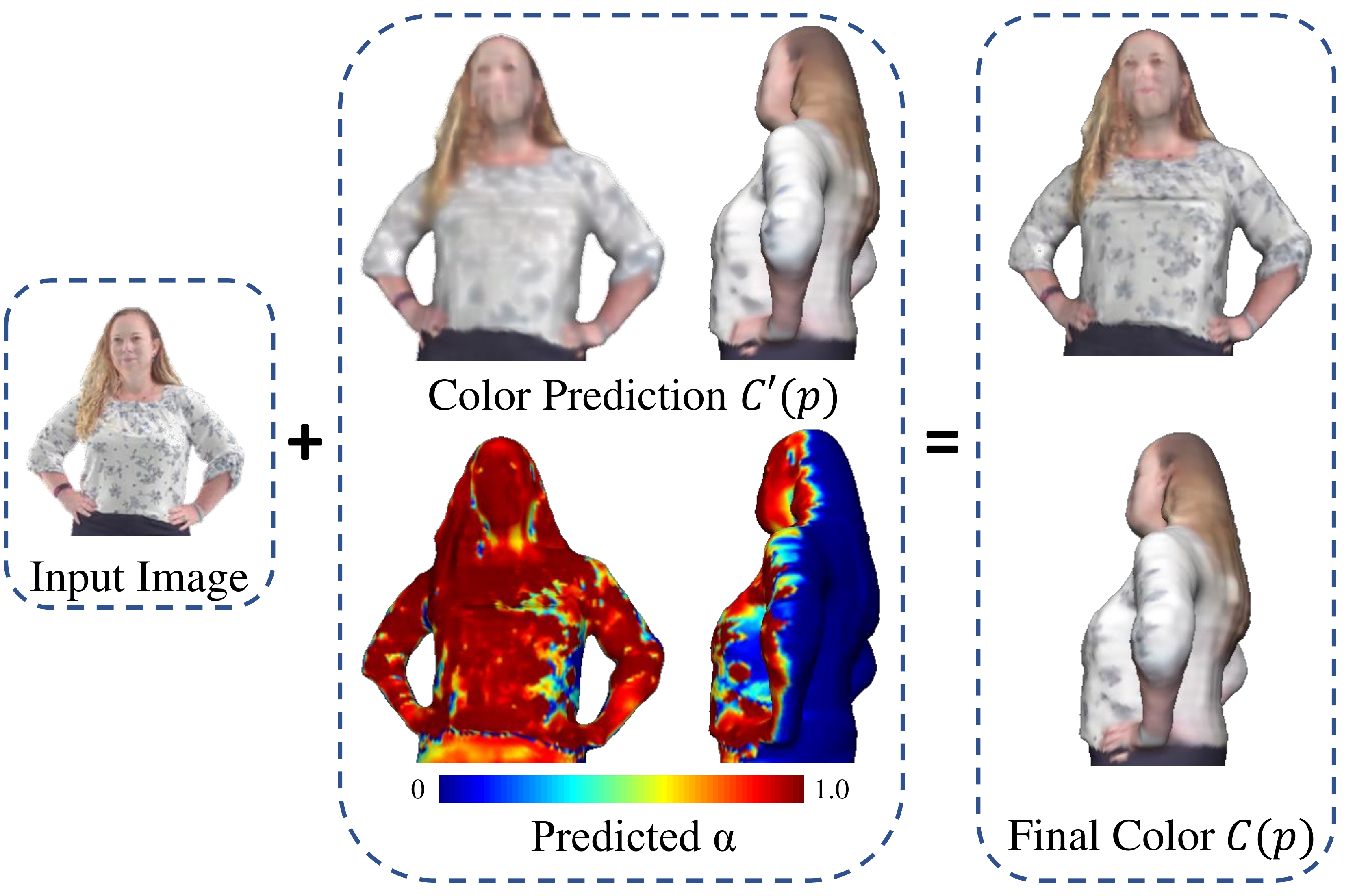}
    \caption{Illustration of our texture inference process. The texture network first recovers the color on the whole surface as well as an alpha channel which is used to blend the predicted color with the observation. In this way, the input image is maximally utilized and more texture details are recovered (Please zoom in to compare the texture on the shirt). }
    \label{fig:vis_attention}
\end{figure}

\subsection{Texture Inference}
\label{sec:method:texture}
Following the practice of Texture Field\cite{texturefield} and PIFu\cite{pifuSHNMKL19}, we also regard the surface texture as a vector function defined in the space near the surface, which can support texturing of shapes with arbitrary topology and self-occlusion. To perform texture inference, we make some simple modification to our network. Specifically, we define the output of the decoder in Fig.\ref{fig:arch} as an RGB$\alpha$ vector field instead of a scalar field. The RGB value is the network prediction of the color of a specific point on the mesh surface, while the alpha channel is used to blend the predicted value with the observed one:
\begin{equation}
    C(p_i) = \alpha * S\left(\mathbf{I}, \pi(p)\right) + (1-\alpha) * C'(p_i)
\end{equation}
where $C'(p_i)$ is the color prediction provided by the network, i.e., the first three channels of the decoder output, $\alpha$ is the last channel of the decoder output and $C(p_i)$ is the final color prediction.
Thus, the reconstruction loss in Eqn.\ref{eqn:reconstruction_loss} is modified accordingly to: 
\begin{equation}
\begin{split}
    \mathcal{L}_R = \frac{1}{n_p}\sum_{i=1}^{n_p}  &\left| C(p_i+ \Delta p_i, c(p_i+ \Delta p_i)) - C^{*}(p_i) \right| +\\ &\left| C'(p_i+ \Delta p_i, c(p_i+ \Delta p_i)) - C^{*}(p_i) \right|, 
\end{split}
\end{equation}
where $C^{*}(p_i)$ is the ground-truth vertex texture and L1 loss is used to avoid color over smoothing. Intuitively, the network learns to infer the color of the whole surface and also determine which part of the surface is visible so that we can directly sample color observation on the image. We show the effect of the alpha channel in Fig.\ref{fig:vis_attention}.

Note that we do not decompose shading component from the original color observation to obtain the ``real'' surface texture. To obtain shading-free texture, one can use state-of-the-art albedo estimation method like \cite{kanamori2018relighting} to firstly remove shading from the input images and then feed the processed images to our texture inference network.

\begin{figure}
	\centering
	\includegraphics[width=\linewidth]{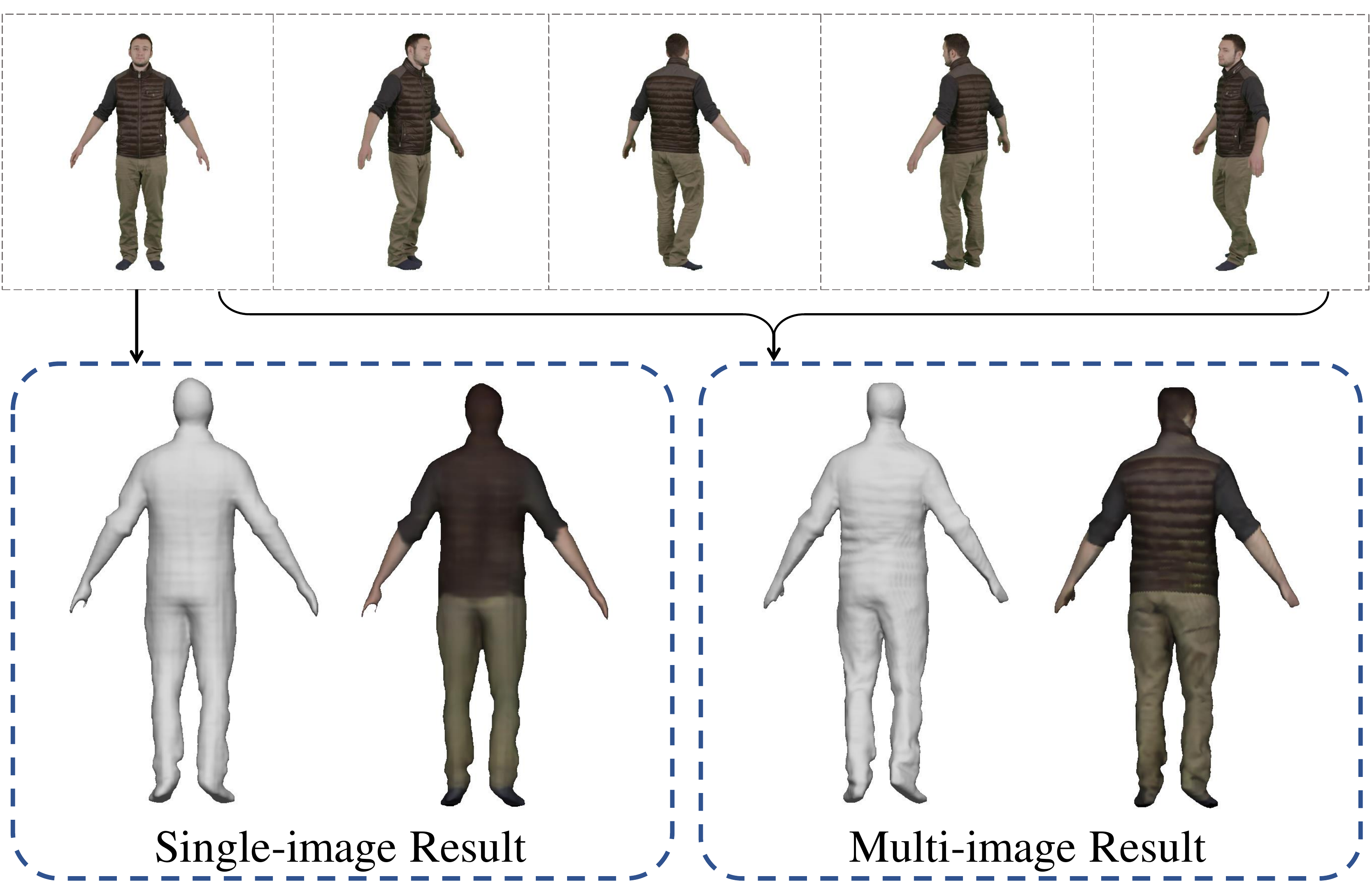}
	\caption{Comparison between the single-image result and the multi-image result. By adding four more frames (without calibration and synchronization infomation) as input, our method is able to recover the surface details on the back. In contrast, the back area is over-smoothed in the single-image setting. Zoom in for better view. }
	\label{fig:eval_multiimage}
\end{figure}

\begin{figure*}
	\begin{center}
		\includegraphics[width=1.0\linewidth]{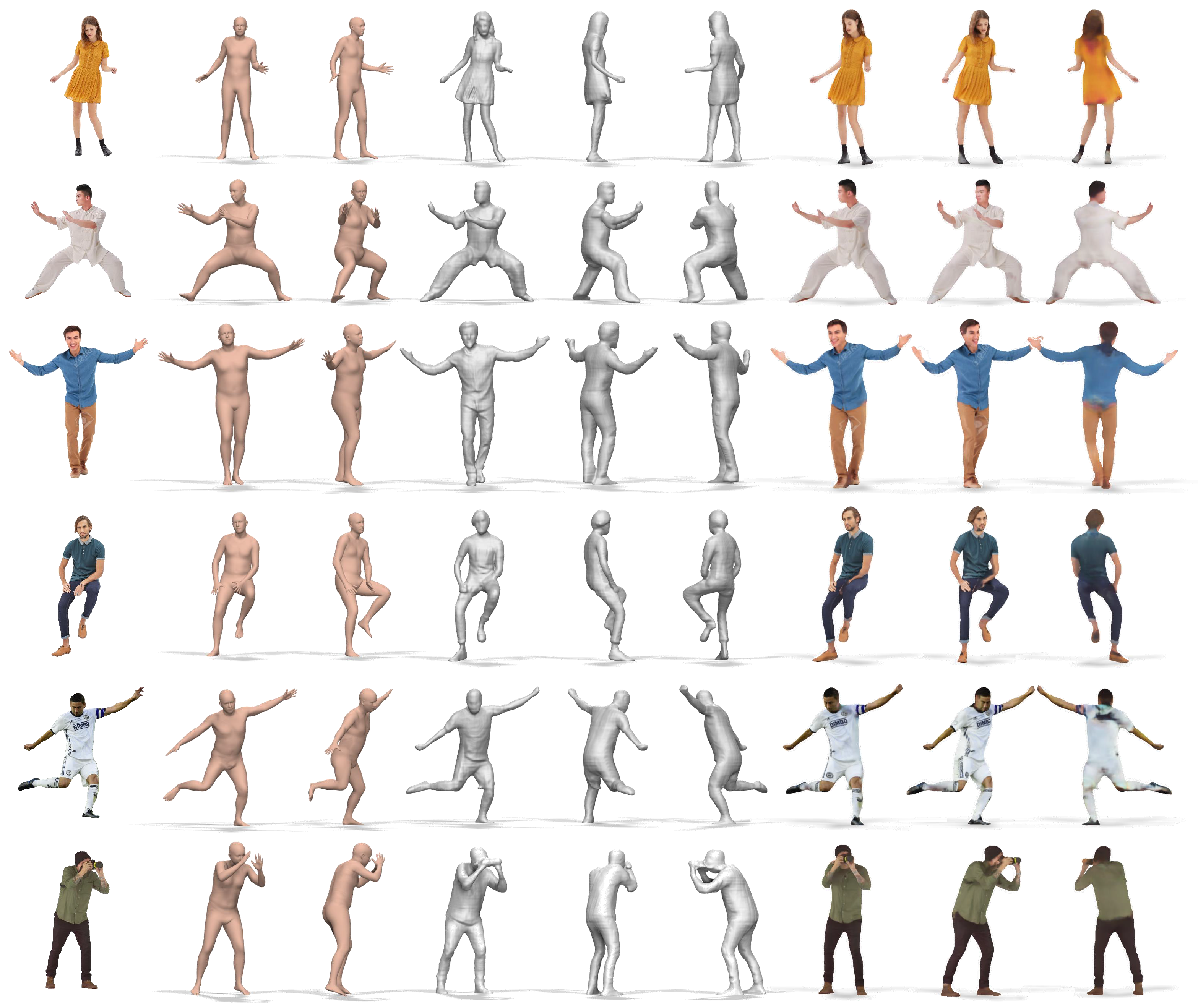}
	\end{center}
	\caption{\revise{Our results on natural images. From left to right: the 1st is the input images, the 2nd and 3rd column show the SMPL models estimated by our method (network inference + optimization), the 4th to 6th demonstrates our geometry reconstruction results, and the last three column demonstrates the texture inference results. The results demonstrate the ability of our method to reconstruct high-quality models and its robust performance to tackle various human poses and clothing styles.}}
	\label{fig:results}
\end{figure*}

\begin{figure*}
	\begin{center}
		\includegraphics[width=1.0\linewidth]{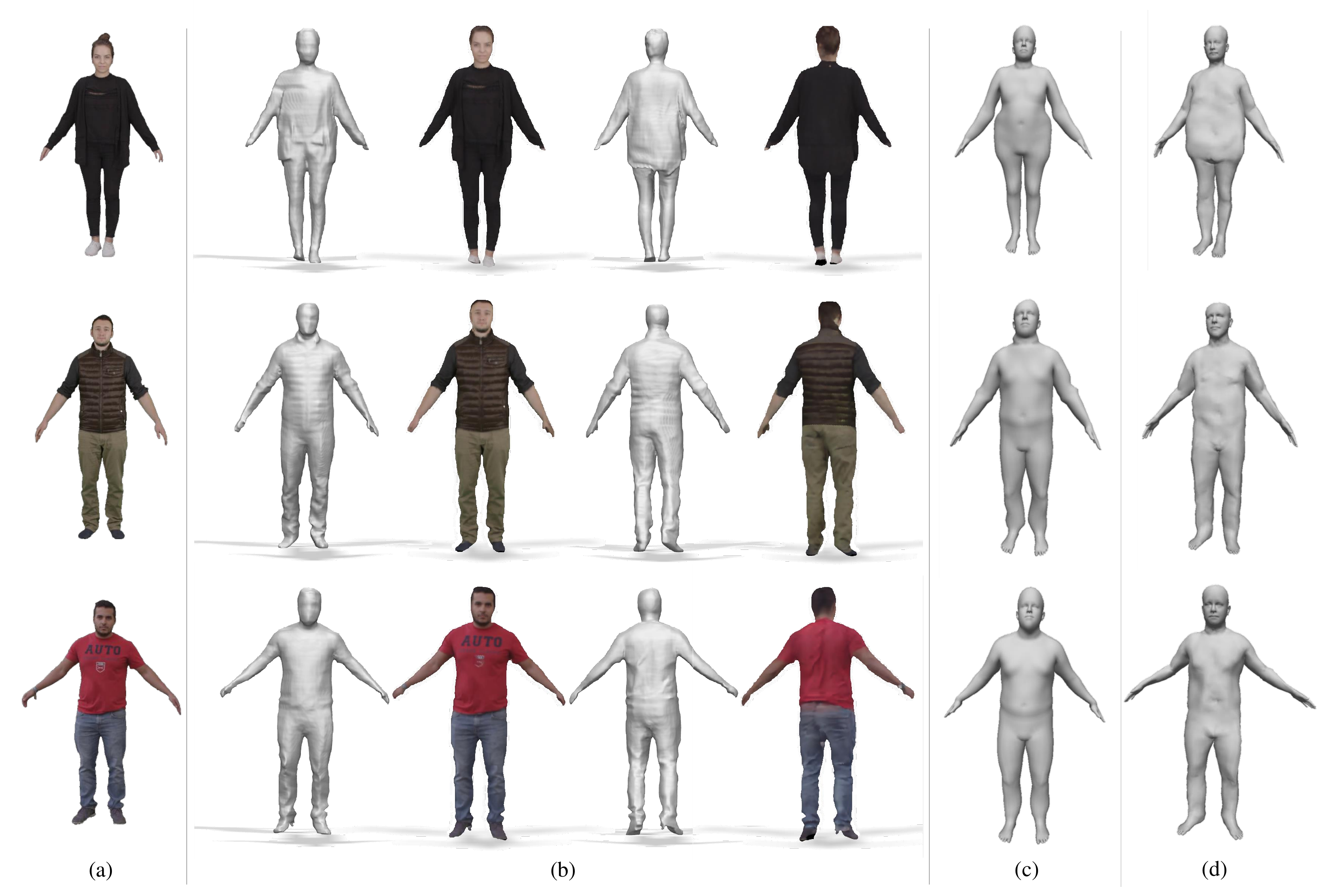}
	\end{center}
	\caption{Multi-image human model reconstruction on VideoAvatar dataset: (a) the first column is the first images of the input sequences, (b) the reconstructed models by our method, (c) results by \cite{VideoAvater}, (d) Results by \cite{alldieck19cvpr}. Our method can reconstruct full-body models with high-resolution details, proving the capability of our PaMIR-based reconstruction method. }
	\label{fig:video_avatar_results}
\end{figure*}

\section{Extension to Multi-image Setup}
\label{sec:method:multiimage}
With the common underlying SMPL model as body reference, our PaMIR representation also implicitly builds a correspondence relationship across different models. As a result, our method can be easily extended to multi-image settings like multi-view input or video input. Specifically, given $N_I$ images of the same person $\mathbf{I}_t, t=0, 1,..., N_I-1$, we first use the optimization method described in Sec.\ref{sec:method:optm} to obtain the SMPL models aligned to each input images, $M(\theta_t, \beta_t)$. Note that the optimization step is performed on each individual image independently; for future work more constraints like shape consistency and pose consistency constraints can be incorporated to the energy function. With these SMPL template parameters we can establish correspondences across two images. For example, for a point $p_i$ in the model space of the reference image $\mathbf{I}_0$, its corresponding point $p_i^{(t)}$ in the model space of $\mathbf{I}_t$ can be calculated as 
\begin{equation}
\begin{gathered}
    p_i^{(t)} = \mathbf{M}_i^{0\rightarrow t} p_i, \\
    \mathbf{M}_i^{0\rightarrow t} = \sum_{j\in\mathcal{N}(p_i)} \frac{\omega_{j\rightarrow i}}{\omega_i} \mathbf{M}_j^{(t)}(\theta_t) \left(\mathbf{M}_j^{(0)}(\theta_0)\right)^{-1}
\end{gathered}
\end{equation}
where $\mathbf{M}_j^{(0)}(\theta_0)$ and $\mathbf{M}_j^{(t)}(\theta_t)$ are the linear blending skinning (LBS) matrices of the $j$-th SMPL vertex in the reference frame and the $t$-th frame, respectively,  $\mathcal{N}(p_i)$, $\omega_{j\rightarrow i}$ and $\omega_i$  share the same definition as Eqn.(\ref{eqn:z_translation}).

To perform geometry inference using multiple image frames, we follow the practice of PIFu\cite{pifuSHNMKL19} and decompose the feature-to-occupancy network $F(\cdot)$ in Eqn.\ref{eqn:pifu} into two components:
\begin{equation}
    F = F_2 \circ F_1
\end{equation}
where $F_1$ is a feature embedding network while $F_2$ is an occupancy reasoning network. For a 3D point $p_i$ in the reference frame $\mathbf{I}_0$, we first sample its feature vector $C^{(0)}(p_i)$  as described in Sec.\ref{sec:surface_representation:pair}. Then we calculate the corresponding point $p_i^{(t)}$ in the $t$-th frame and also sample its corresponding feature vector $C^{(t)}(p_i^{(t)})$, where $t=1, 2, ..., N_I-1$. The feature embedding network $F_1$ takes these feature vectors and encodes them into latent feature embedding vectors $\Phi_i^{(k)}, k=0, 1, ..., N_I-1$. The latent embedding vectors across different frames are aggregated into one embedding $\bar{\Phi}_i$ using mean pooling, which is then mapped to the target implicit field $F(p_i)$ by the occupancy reasoning network  $F_2$. The multi-image network is fine-tuned from the single-image network using multi-view rendering of 3D human models. 

Note that PIFu also demonstrates results given mutli-view inputs. However, the multi-view input should be well calibrated and synchronized. In contrast, neither calibration or synchronization is necessary in our method because we can utilize the SMPL estimation to build the correspondence across different views. Moreover, our method can handle the cases where body poses are not identical across images, e.g., video input. This feature enables video-based personalized avatar creation. A similar application has been demonstrated in \cite{VideoAvater}, but our method can support more challenging cloth topologies and recover more surface details. Overall, in terms of multi-image setup, our method is more general and practical than state-of-the-art methods. A comparison between the single-image result and the multi-image result is presented in Fig.\ref{fig:eval_multiimage}. As shown in the figure, by adding four more video frames as input, our method is able to recover the surface details on the back that is invisible in the first frame.

Since we consider only pose changes across views and neglect surface detail deformations (e.g., wrinkle movements) across frames, challenging pose deviation and inconsistent geometry between frames will cause inaccurate feature fusion and reconstruction artifacts. Besides, the multi-image network is trained using only multi-view rendering of human models without pose deviations. For future work, we can introduce 4D data to resolve these issues.

\begin{figure*}
    \centering
    \includegraphics[width=0.48\linewidth]{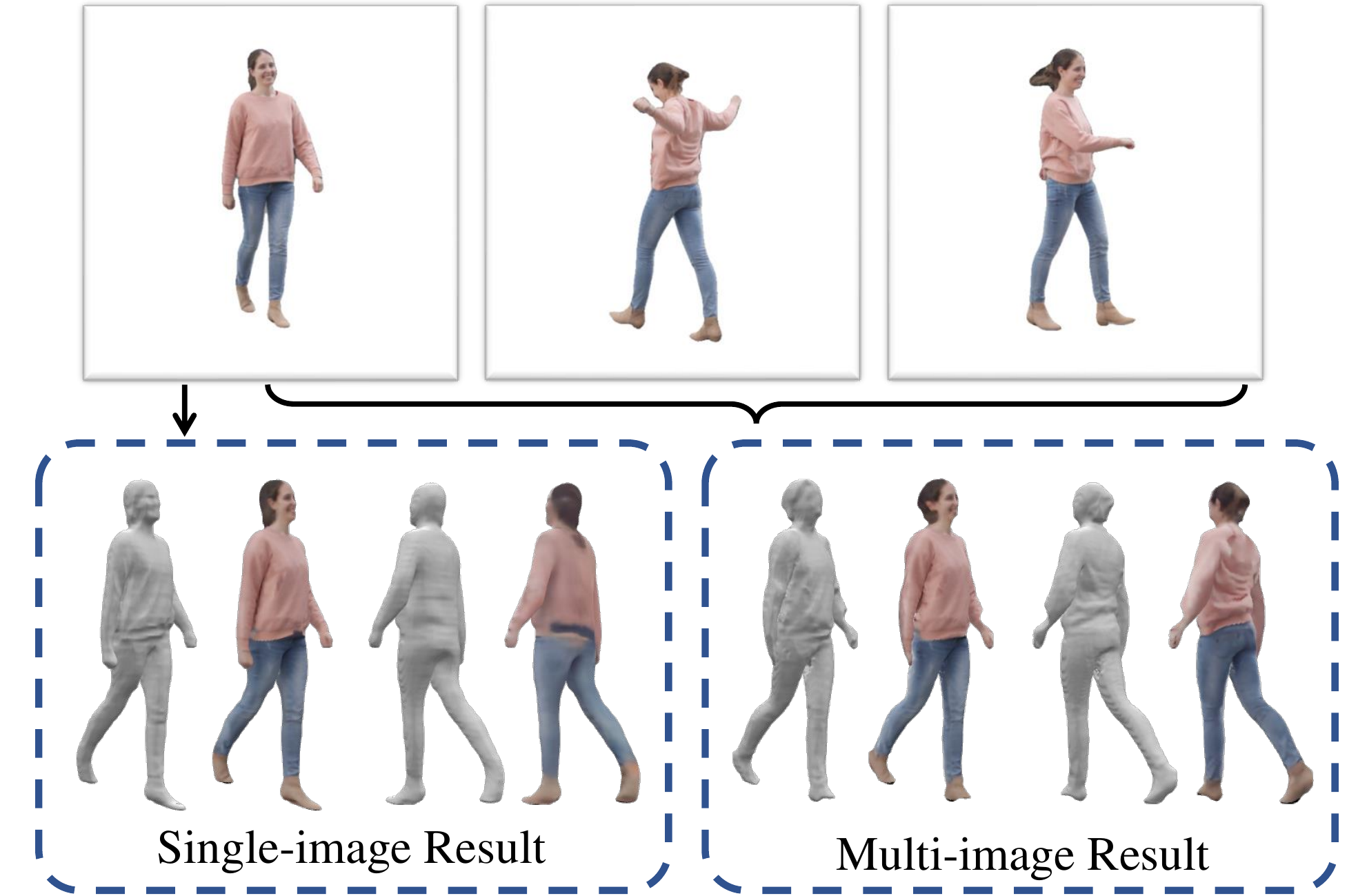}
    \includegraphics[width=0.48\linewidth]{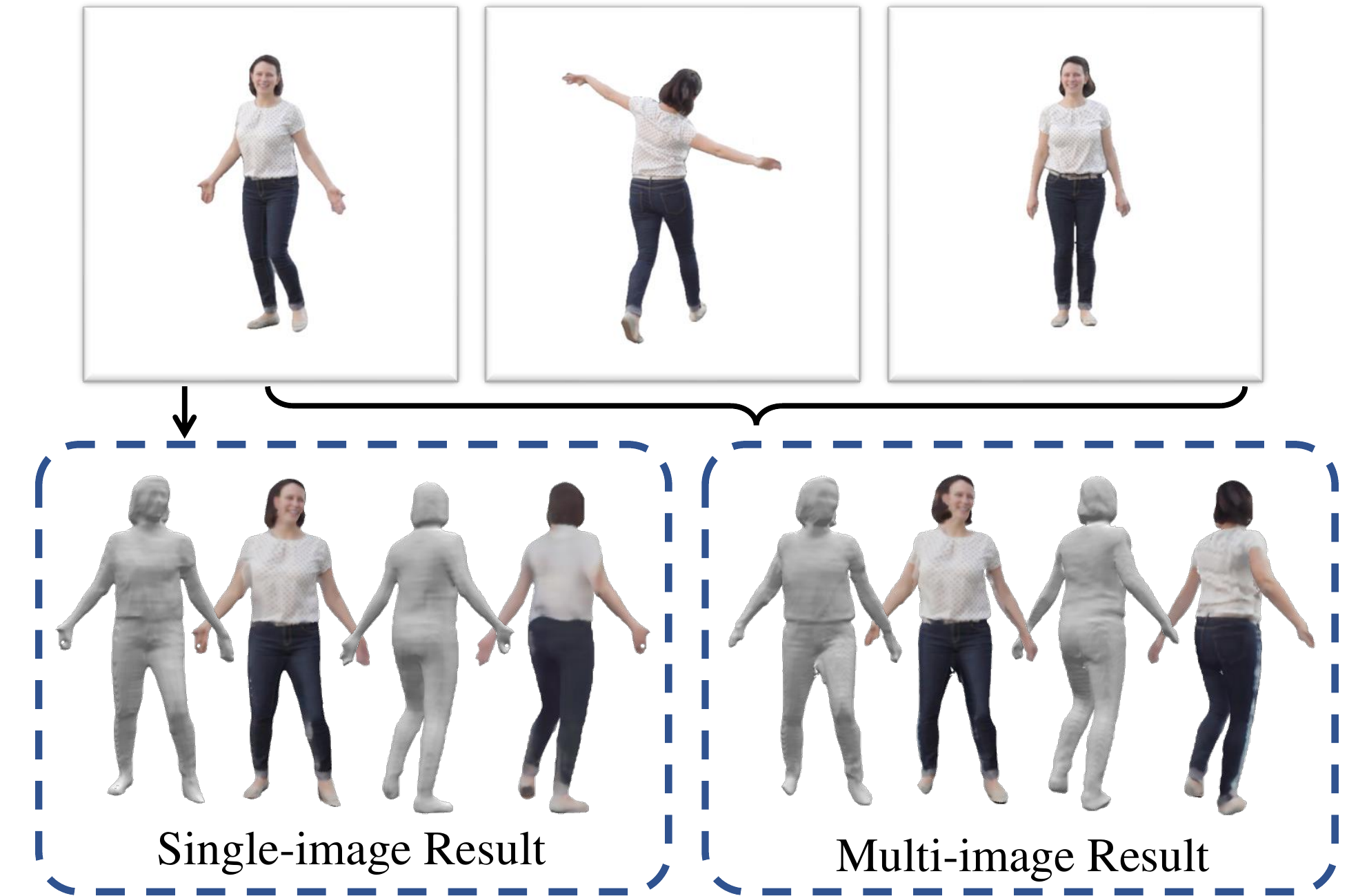}
    \caption{Comparison between the single-image reconstruction results and the multi-image results from input images with moderate pose deviations. In these  cases, our method is still able to integrate information from different frames.}
    \label{fig:challenging_multi_images}
\end{figure*}

\section{Experiments}
In this section we evaluate our approach. Details about the implementation and the evaluation dataset are given in Sec.\ref{sec:results:impl_details} and Sec.\ref{sec:results:dataset} respectively. In Sec.\ref{sec:results:results} we demonstrate that our method is able to reconstruct human models with challenging poses. We then compare our method against state-of-the-art methods in Sec.\ref{sec:results:comparison}. We also evaluate our contributions both qualitatively and quantitatively in Sec.\ref{sec:results:ablation}. The quantitative evaluation results are given in Tab.\ref{tab:quant_evalation}. We mainly use geometry reconstruction performance for evaluation, which is our main focus. 
\label{sec:results}

\subsection{Implementation Details}
\label{sec:results:impl_details}

\textbf{Network Architecture. } 
For image feature extraction, we adapt the same 2D image encoders in PIFu\cite{pifuSHNMKL19} (i.e., Hourglass Stack\cite{Hourglass16} for geometry and CycleGAN\cite{CycleGAN2017} for texture), which take as input an image of 512$\times$512 resolution and outputs a 256-channel feature map with a resolution of 128$\times$128. For volumetric feature extraction, we use a 3D convolution network which consists of two convolution layers and three residual blocks. Its input resolution is 128$\times$128$\times$128 and its output is a 32-channel feature volume with a resolution of 32$\times$32$\times$32. We replace batch normalization with group normalization to improve the training stability. The  feature decoder is implemented as a multi-layer perceptron, where the number of neurons is $(288, 1024, 512, 256, 128, ch)$, where $ch=1$ for the geometry network while  $ch=4$ for the texture network. 

\textbf{Training Data. } 
To achieve state-of-the-art reconstruction quality, we collect 1000 high-quality textured human scans with various clothing, shapes, and poses from Twindom\footnote{\url{https://web.twindom.com/}}. We randomly split the Twindom dataset into a training set of 900 scans and a testing of 100 scans. To augement pose variety, we also randomly sample 600 models from DeepHuman\cite{Zheng2019DeepHuman} dataset. 
We render the training models from multiple view points using Lambertian diffuse shading and spherical harmonics lighting\cite{SURREAL}. We render the images with a weak perspective camera and image resolution of 512 $\times$ 512. To obtain the ground-truth SMPL annotations for the training data, we apply MuVS\cite{huang2017muvs} to the multi-view images for pose computation and then solve for the shape parameters to further register the SMPL to the scans. For point sampling during training, we use the same scheme proposed in PIFu\cite{pifuSHNMKL19}, in which the authors combine uniform sampling and adaptive sampling based on the surface geometry and use embree algorithm\cite{wald2014embree} for occupancy querying. 

\textbf{Network training. } 
We use Adam optimizer for network training with the learning rate of $1\times 10^{-3}$, the batch size of 3, the number of epochs of 10, and the number of sampled points of 5000 per subject. The learning rate is decayed by the factor of 0.1 at every 10000-th iteration.
We also combine predicted and ground-truth SMPL models during network training. Specifically, in each batch, we randomly select one image to replace the predicted SMPL model with the ground-truth when constructing the depth-ambiguity-aware reconstruction loss. In this way we can guarantee the best performance is obtained once the underlining SMPL becomes more accurate. 
The multi-image network is fine-tuned from the models trained for single-view network with three random views of the same subject using a learning rate of $2\times 10^{-5}$ and a batch size of 1. Before training the PaMIR network, we first fine-tune the pre-trained GCMR network on our training set. 

\textbf{Network testing. } 
When testing, our network only requires an RGB image as input and outputs both the parametric model and the reconstructed surface with texture. To maximize the performance, We run the body reference optimization step for all results  unless otherwise stated. Fifty iterations are needed for the optimization and take about 40 seconds. 

\textbf{Network complexity. } 
We report the module complexity in Tab.\ref{tab:complexity}. To reconstruct the 3D human model given an RGB image, we use the hierarchical SDF querying method proposed in OccNet\cite{Mescheder2019OccupancyNetwork} to reduce network querying times. Overall, taking the body reference optimization step into account, it takes about 50s to reconstruct the geometry of the 3D human model and 1s to recover its surface color. 

\begin{table}
    \centering
    \caption{The number of parameters and execution time of each network module}
    \label{tab:complexity}
	\begin{threeparttable}
    \begin{tabular}{lcc}
        \toprule
         Module & \#Parameters & Execution Time  \\
         \midrule
         GCMR  & 46,874,690 & 0.15s\tnote{$\ast$} \\
         Geometry Network  & 27,225,105 & 0.25s\tnote{$\dag$}\\ 
         Texture Network & 13,088,268 & 0.29s\tnote{$\dag$}\\
         \bottomrule
    \end{tabular}
	\begin{tablenotes}
        \scriptsize
        \item[$\ast$] Measured using one RGB image. 
        \item[$\dag$] Measured using one RGB image and 10k query points. 
        \end{tablenotes}
        \end{threeparttable}
 \end{table}

\begin{figure*}
  \centering
  \includegraphics[width=1.0\linewidth]{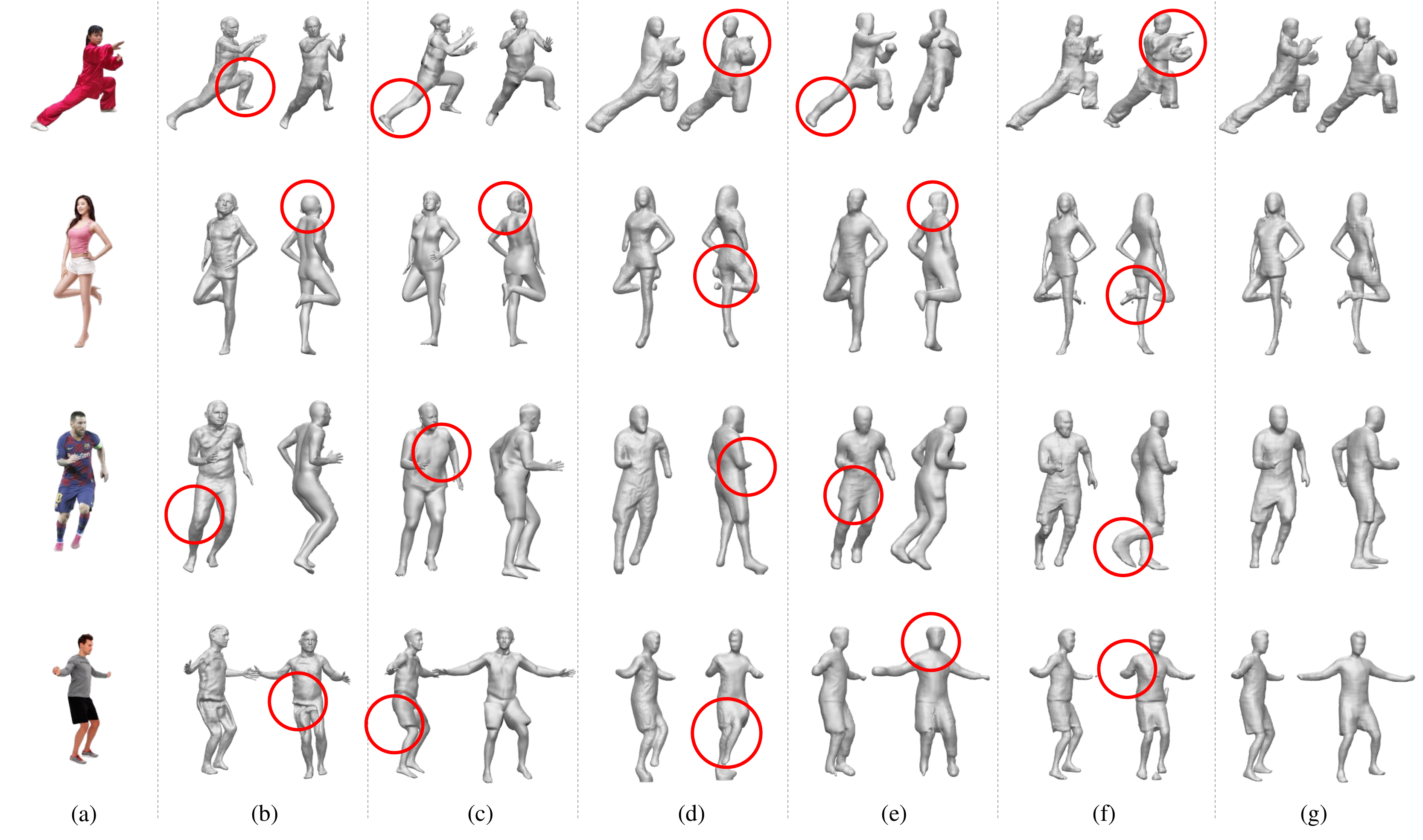}
  \caption{Qualitative comparison against state-of-the-art methods for single-image human model reconstruction: (a) input images,  (b) results by HMD\cite{HMD2019}, (c) Tex2Shape\cite{tex2shape2019}, (d) Moulding Humans\cite{MouldingHumans2019}, (e) DeepHuman\cite{Zheng2019DeepHuman}, (f) PIFu\cite{pifuSHNMKL19}, (g) ours. }
  \label{fig:comparison} 
\end{figure*}

\subsection{Evaluation Dataset}
\label{sec:results:dataset}
For qualitative evaluation on single-image reconstruction, we utilize real-world full-body images collected from the DeepFashion dataset\cite{liuLQWTcvpr16DeepFashion} and from the Internet. We remove the background of these real-world image using neural semantic segmentation\cite{liang2018LIP} followed by Grabcut refinement\cite{rother2004grabcut}. For qualitative evaluation on multi-image reconstruction, we use the open-source dataset from VideoAvatar\cite{VideoAvater} which contains 24 RGB sequences of different subjects turning 360 degree with a rough A-pose in front of a camera. The testing split of the Twindom dataset is used for quantitative comparison. In addition, for quantitative comparison and evaluation we also use the BUFF dataset\cite{shapeundercloth:CVPR17}, which provides 26 4D human sequences with various clothes and over 9000 scans in total. To reduce computation burden, We select 300 scans of different poses from the buff dataset. We render both the Twindom testing data and the BUFF data from 12 views spanning every 30 degrees in yaw axis using the same method in Sec.\ref{sec:results:impl_details}.


\subsection{Results}
\label{sec:results:results}

We demonstrate our approach for single-image 3D human reconstruction in Fig.\ref{fig:teaser} and Fig.\ref{fig:results}. The input images in Fig.\ref{fig:teaser} and Fig.\ref{fig:results} covers various body poses (dancing, Kungfu, sitting and running), and also covers different clothes (loose pants, skirts, sports suits and casual clothes). The results demonstrate the ability of our method to reconstruct high-quality 3D human models and its robust performance to tackle various human poses and clothing styles. 

We also test our performance for multi-image human model reconstruction using the VideoAvatar dataset in Fig.\ref{fig:video_avatar_results}. We uniformly sample 5 frames for each sequence for surface reconstruction. Note that for each subject, the poses in different frames are different due to the body movements and no camera extrinsic parameter is provided, so PIFu is not suitable to reconstruct human models in this case. In contrast, our method can reconstruct full-body models with high-resolution details, proving the generalization capability of our PaMIR representation. 
We also present two example results of applying multi-image feature fusion on images with moderate pose deviation in Fig.\ref{fig:challenging_multi_images}. As the figure shows, our method is still able to reconstruct the overall shapes of the subjects in these cases.

\subsection{Comparison}
\label{sec:results:comparison}

We qualitatively compare our method with several state-of-the-art methods including HMD\cite{HMD2019}, Tex2Shape\cite{tex2shape2019}, Moulding Humans\cite{MouldingHumans2019}, DeepHuman\cite{Zheng2019DeepHuman} and PIFu\cite{pifuSHNMKL19}. Among them, HMD\cite{HMD2019} and Tex2Shape\cite{tex2shape2019} are parametric methods based on SMPL\cite{SMPL:2015} model deformation, PIFu\cite{pifuSHNMKL19} uses a deep implicit function as geometry representation, Moulding Humans\cite{MouldingHumans2019} uses the combination of a front depth map and a back depth map for representation, and DeepHuman\cite{Zheng2019DeepHuman} combines volumetric representation with the SMPL model. Note that Tex2Shape only computes shapes, so we use the pose parameters obtained by our method to skin the its results. 
For simplicity, we omit comparisons with the works that have already been compared like BodyNet\cite{BodyNet} and SiCloPe\cite{SiCloPe2019}.
In our experiments, DeepHuman, PIFu and Moulding Humans are all retrained on the our dataset, while parametric methods like HMD and Tex2shape are not because our dataset contains loose garments like dresses which will deteriorate the performance of those methods. 



We conduct qualitative comparison in Fig.\ref{fig:comparison}. As shown in the figure, HMD and Tex2Shape have difficulties dealing with loose clothes and cannot reconstruct the surface geometry accurately; Moulding Human\cite{MouldingHumans2019} fails to handle challenging poses (due to the lack of semantic constraints), and produces broken body parts when self-occlusions occur, which is the essential limitation of its double depth representation;  DeepHuman\cite{Zheng2019DeepHuman} cannot recover high-frequency geometrical details although it succeeds to reconstruct the rough shapes from the images;  PIFu\cite{pifuSHNMKL19} struggles to reconstruct the models in challenging poses and also suffer from self-occlusions. In contrast, our method is able to reconstruct plausible 3D human models under challenging body poses and various clothing styles. In terms of surface quality and pose generalization capacity, our method is superior to other state-of-the-art methods. 


\begin{figure*}
	\begin{center}
		\includegraphics[width=0.9\linewidth]{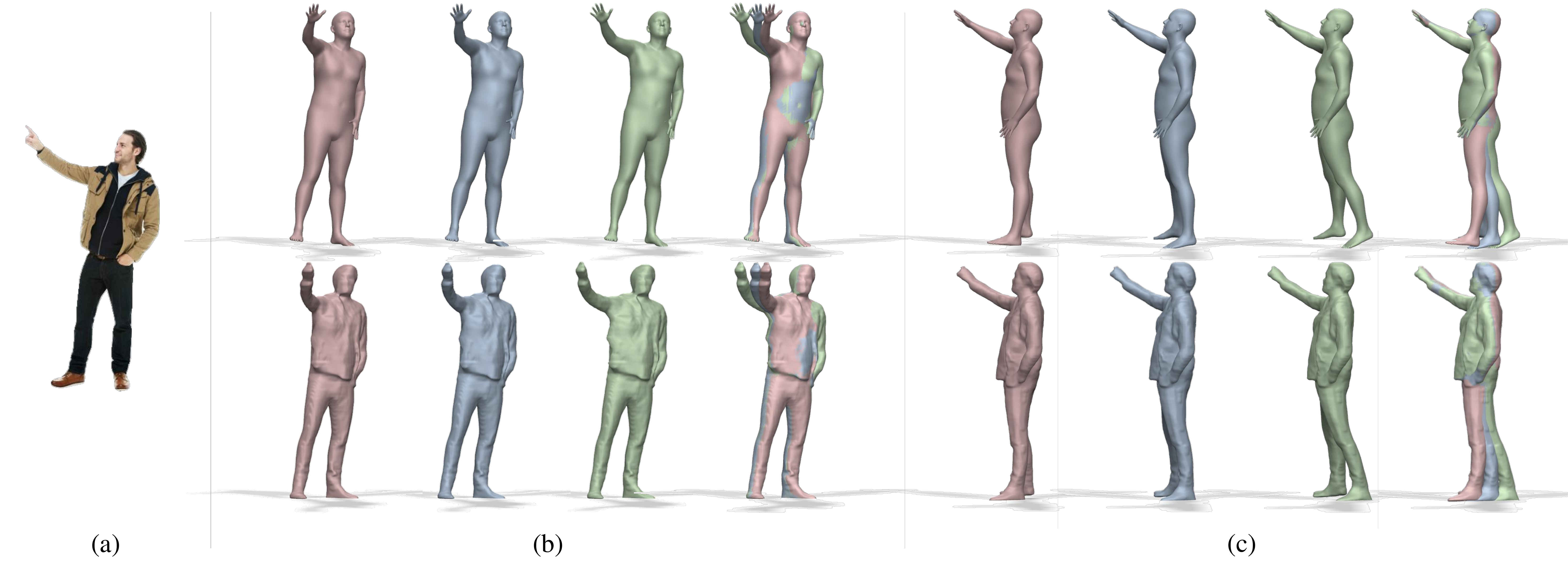}
	\end{center}
	\caption{Two examples of multi-modal output. Given an input image (a), our method can output different possible reconstruction results (top row of (b)(c)) corresponding to different body pose hypothesis (bottom row of (b)(c) ). From the overlapped figure (the last columns of (b)(c)), we can see that the right arm and the left leg of reconstructed mesh adjusts accordingly with the right arm and the left leg of the body hypothesis, while other parts of the mesh keep consistent. Better view in color. }
	\label{fig:pair1}
\end{figure*}

\begin{table*}
	\caption{Numerical comparison results.  }
	\centering
	\begin{tabular}{l|c|c|c|c}
        \hline
        Dataset & \multicolumn{2}{c}{BUFF} & \multicolumn{2}{|c}{TWINDOM} \\
        \hline
		Method & P2S (cm) &  Chamfer (cm) & P2S (cm) &  Chamfer (cm)  \\
		\hline
		HMD\cite{Zheng2019DeepHuman} & 2.48 & 3.92 & 2.50 & 4.01\\
		Moulding Humans\cite{MouldingHumans2019} & 2.25 & 2.68 & 2.84 & 3.35 \\
		DeepHuman\cite{Zheng2019DeepHuman} & 2.15 & 2.80 & 2.35 & 2.97\\
		PIFu\cite{pifuSHNMKL19} & 1.93 & 2.22 & 2.34 & 2.65\\
		Ours & 1.52 & 1.92 & 1.80 & 2.12 \\
		Ours using ground-truth SMPL & 0.709 & 0.936 & 0.744 & 1.00 \\
        \hline
	\end{tabular}
	\label{tab:quant_comparison}
\end{table*}

We quantitatively compare our method with the state-of-the-art methods using both Twindom testing dataset and BUFF rendering dataset to evaluate the geometry reconsturction accuracy. 
Similar to the experiments in PIFu\cite{pifuSHNMKL19}, we use point-to-surface error as well as the Chamfer distance as error metric. 
The numerical results are presented in Tab.\ref{tab:quant_comparison}. The quantitative comparison shows that our method outperforms the state-of-the-art methods in terms of surface reconstruction accuracy. 
We also provide the errors when ground-truth SMPL annotations are available to present the upper limit of our reconstruction accuracy if the SMPL estimation is perfect. Overall, our method is more general, more robust and more accurate than HMD\cite{HMD2019},  Moulding Humans\cite{MouldingHumans2019}, DeepHuman\cite{Zheng2019DeepHuman} and PIFu\cite{pifuSHNMKL19}.


For multi-image setups, we also conduct qualitative comparison against state-of-the-art methods \cite{VideoAvater} and \cite{alldieck19cvpr}. 
The method proposed in \cite{VideoAvater} is an optimization-based method which deforms the SMPL template according to the silhouettes.
The optimization is performed on the whole video sequence. 
In contrast, the method in \cite{alldieck19cvpr} is a learning-based method that deforms the SMPL template based on only 8 views of images. 
The comparison results are shown in Fig.\ref{fig:video_avatar_results}. 
From the results we can see that although all methods are able to reconstruct the overall shapes correctly, our method is able to recover more surface details than the other two methods. 
This is because our non-parametric representation allows more flexible surface reconstruction. 
We do not perform quantitative comparison because there is no such benchmark available.

\subsection{Evaluation}
\label{sec:results:ablation}

\subsubsection{PaMIR Representation}

The superiority of our PaMIR representation is already proved through the comparison experiments in Sec.\ref{sec:results:results} and \ref{sec:results:comparison}. In this evaluation, we demonstrate the advantage of our PaMIR representation from another aspect: it can support multi-modal outputs. To be more specific, our method can output multiple possible human models corresponding to different body pose hypotheses. Two examples are shown in Fig.\ref{fig:pair1}. In both experiments we manually adjust one part of the body in order to generate multiple pose hypotheses. As a result, the reconstruction results change in accordance with the input SMPL models. In contrast, non-parametric method like PIFu\cite{pifuSHNMKL19} or Moulding Humans\cite{MouldingHumans2019} can only generate a specific mesh for each image, showing their limited and overfit capability.

\begin{table}
	\caption{Numerical Ablation Study.  }
	\centering
	\begin{threeparttable}
	\begin{tabular}{lc}
		\toprule
		Method\tnote{$\ast$} & P2S (cm)   \\
		\midrule
		GT SMPL & 1.71 \\
		Pred SMPL wo/ DAAL& 1.81 \\
		Pred SMPL w/ DAAL& 1.60\\
		Full (Pred SMPL w/ DAAL + Optimization) &  1.52 \\
		\bottomrule
	\end{tabular}
	\begin{tablenotes}
	   \scriptsize
       \item[$\ast$] GT = ground-truth, Pred = predicted, DAAL = depth-ambiguity-aware loss
     \end{tablenotes}
    \end{threeparttable}
	\label{tab:quant_evalation}
\end{table}

\subsubsection{Depth-ambiguity-aware Loss}

\begin{figure}
	\begin{center}
		\includegraphics[width=1.0\linewidth]{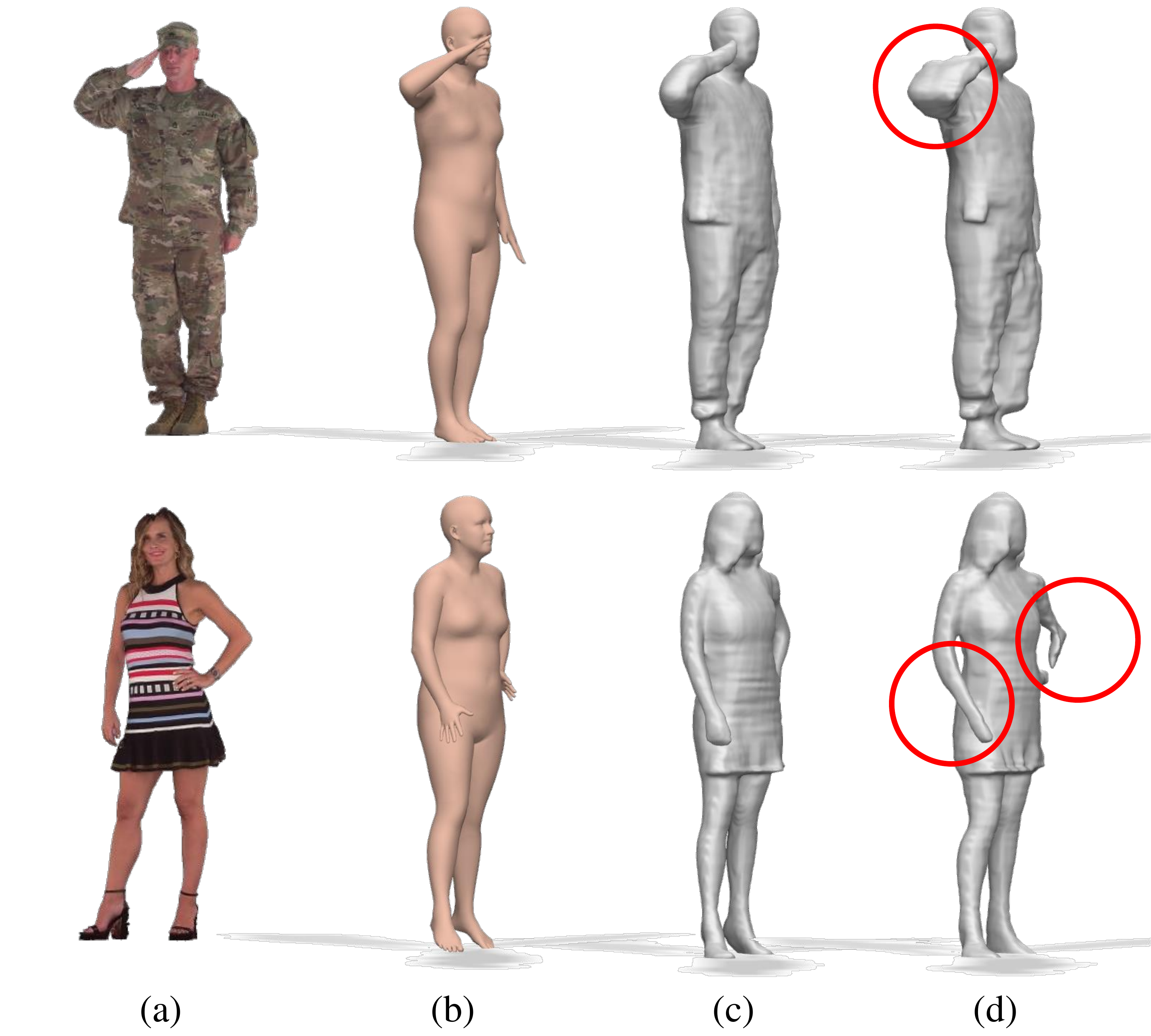}
	\end{center}
	\caption{Qualitative evaluation on depth-ambiguity-aware training loss: (a) the input images, (b) the corresponding SMPL models estimated by GCMR, (c) our reconstruction results, (d) the reconstruction results of the baseline network. }
	\label{fig:eval_adaptive_loss}
\end{figure}

To conduct ablation study on our depth-ambiguity-aware loss, we implement a baseline network that is trained using the traditional reconstruction loss and compare it against our network.  After the same number of training epochs, we test both network on real-world input images. Two example results are shown in Fig.\ref{fig:eval_adaptive_loss}. As we can see in the figure, with our depth-ambiguity-aware reconstruction loss, the network is able to learn to reconstruct 3D human models that have consistent poses with the estimated SMPL models. On the contrary, the models reconstructed by the baseline network without our depth-ambiguity-aware loss are less consistent with the SMPL models and have more artifacts. The quantitative study is presented in Tab.\ref{tab:quant_evalation}. From the numeric comparison between the 3rd and 4th rows of Tab.\ref{tab:quant_evalation}, we can conclude that our  depth-ambiguity-aware loss avoids the negative impact of the depth inconsistency between predicted SMPL models and ground-truth scans, and improves the accuracy of the reconstruction results. 

\subsubsection{Training Scheme}
\label{sec:eval:end2end}

\begin{figure}
	\begin{center}
		\includegraphics[width=1.0\linewidth]{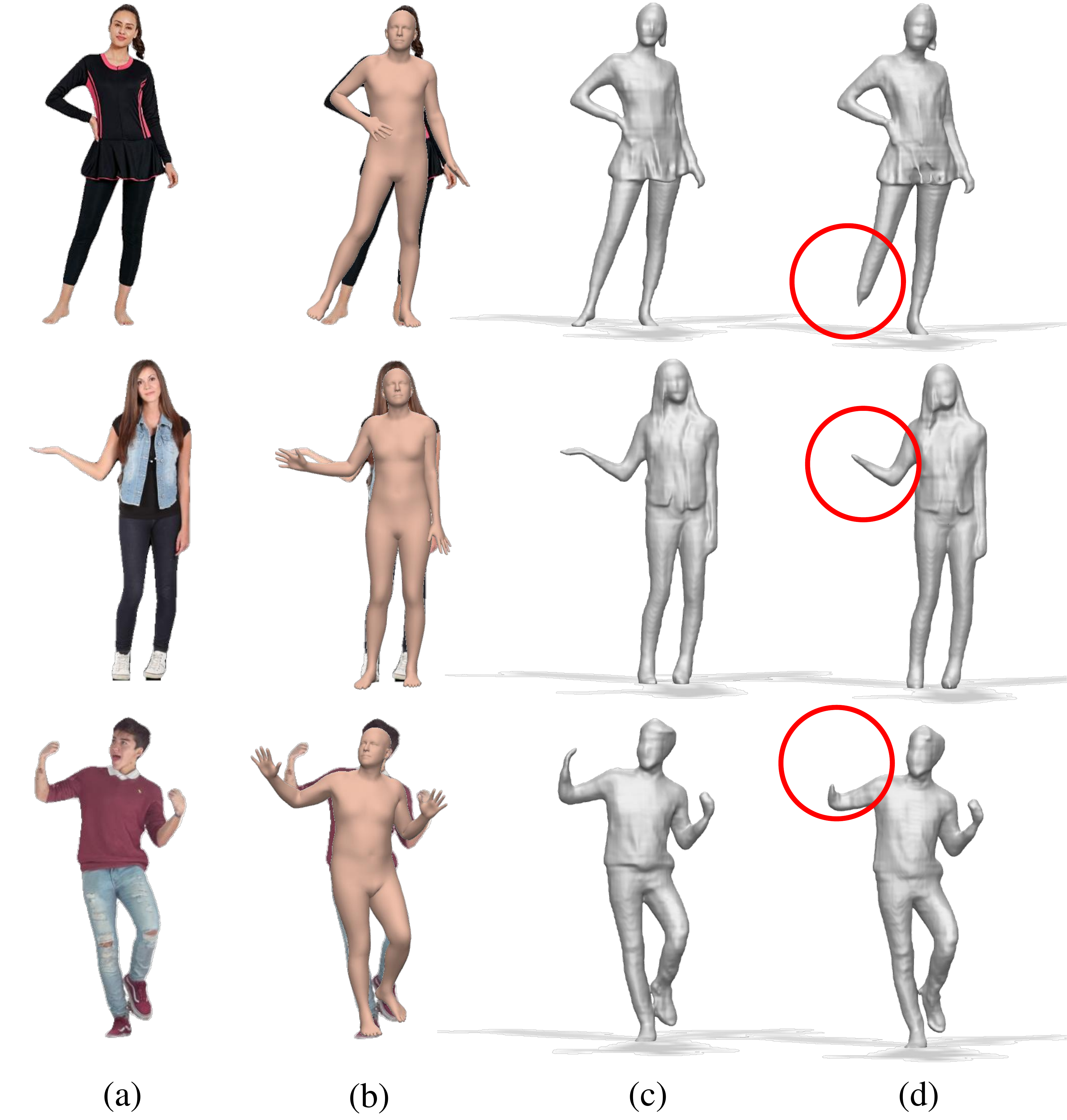}
	\end{center}
	\caption{Qualitative evaluation on our training scheme: (a) the input images, (b) the estimated SMPL models rendered on top of the input images, (c) our reconstruction results, (d) the reconstruction results of the baseline network. }
	\label{fig:eval_endtoend}
\end{figure}

We also implement another baseline network that is trained using the ground-truth SMPL annotations to evaluate our training scheme. We compare this baseline network with our network on different input images. In Fig.\ref{fig:eval_endtoend}, we present some cases in which the predicted SMPL models are not well aligned with image keypoints and silhouettes. As we can see in Fig.\ref{fig:eval_endtoend}, under the scenarios of inaccurate SMPL estimation, the baseline network fails to reconstruct complete full-body models, while our network is able to provide plausible results in accordance with image observations. Note that this feature is of vital importance. For instance, in the last example of Fig.\ref{fig:eval_endtoend}, the right forearm of the baseline model is completely missing. In this case, the reference body optimization step could not work because no matter how the right forearm moves, the occupancy probability value of its vertices are always closed to zero. The quantitative study is presented in Tab.\ref{tab:quant_evalation}. From the numerical comparison between the 2nd and 4th rows of Tab.\ref{tab:quant_evalation}, we can conclude that our training scheme improves the accuracy of our reconstruction results at inference time when no accurate SMPL annotation is available. 

\subsubsection{Reference Body Optimization}

\begin{table}
    \centering
    \caption{Mean Per Joint Position Error (MPJPE, unit: cm) Before/After Body Reference Optimization. }
    \label{tab:smpl_accu}
    \begin{tabular}{lcc}
        \toprule
         Dataset & BUFF & TWINDOM   \\
         \midrule
         Before Optimization & 2.65 & 2.85 \\
         After Optimization  & 2.49 & 2.74 \\ 
         \bottomrule
    \end{tabular}
 \end{table}

\begin{figure}
	\begin{center}
		\includegraphics[width=1.0\linewidth]{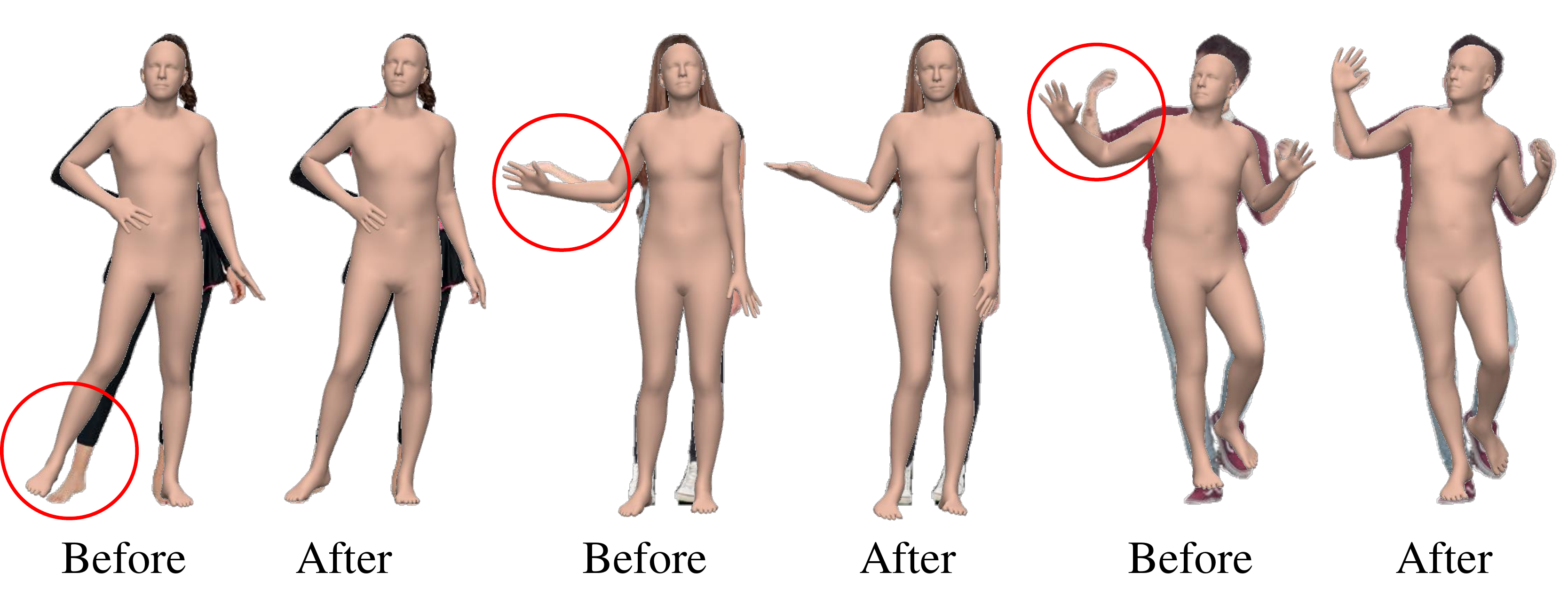}
	\end{center}
	\caption{Evaluation of reference body optimization. Note that after body reference optimization, the SMPL models are more accurately aligned with the image observations. }
	\label{fig:eval_body_optimization}
\end{figure}

To evaluate the effectiveness of our reference body optimization step, we compare the body fitting results before and after optimization using the evaluation images in Sec.\ref{sec:eval:end2end}. The results are presented in Fig.\ref{fig:eval_body_optimization}. As shown in the figure, the optimization step can further register the SMPL model to the image observation, resulting into more accurate body pose estimation. This is also proven in the quantitiative evaluation in Tab.\ref{tab:smpl_accu}.  From the numerical results in the last two rows of Tab.\ref{tab:quant_evalation}, we can also see that the mesh reconstruction is also improved after reference body optimization.

\subsubsection{Multi-image Setup}
To evaluate the detail changes using more or less images, we conduct a quantitative evaluation in Tab.\ref{tab:normal_improvement}. Specifically, we perform feature fusion with 1, 2, 3 and 4 views of the same model and extract the meshes. Then we measure the normal reprojection error\cite{pifuSHNMKL19} from 8 uniform viewpoints to evaluate the detail improvements. \revise{The numerical results presented in Tab.\ref{tab:normal_improvement}  suggest that more geometric details are recovered after fusing information from multiple input images. }

\begin{table}
    \centering
    \caption{Normal Reprojection Error with Inputs Images From Various Numbers of View Points. }
    \label{tab:normal_improvement}
    \begin{tabular}{lcccc}
        \toprule
         View Num & 1 & 2 & 3 & 4   \\
         \midrule
         Normal Reprojection Error & 0.161 & 0.147 & 0.139 & 0.135 \\
         \bottomrule
    \end{tabular}
 \end{table}



\section{Discussion}
\noindent\textbf{Conclusion. }
Modeling 3D humans accurately and robustly from a single RGB image is an extremely ill-posed problem due to the varieties of body poses, clothing types, view points and other environment factors. Our key idea to overcome these challenges is factoring out pose estimation from surface reconstruction. To this end, we have contributed a deep learning-based framework to combine the parametric SMPL model and the non-parametric deep implicit function for 3D human model reconstruction from a single RGB image. Benefiting from the proposed PaMIR representation, the depth-ambiguity-aware reconstruction loss and the reference body optimization algorithm, our method outperforms state-of-the-art methods in terms of both robustness and surface details.  As shown in the supplementary video, we can obtain temporally consistent reconstruction results by applying our method to video frames individually. We believe that our method will enable many applications and further stimulate the research in this area. 

From a more general perspective of view, we have made a step forward towards integrating semantic information into the free-form implicit fields which have attracted more and more attention from the research community for its flexibility, representation power and compact nature. For example, similar topics include the reconstruction of hand and face using such PaMIR representation. We also believe our semantic implicit fields would become one important future topic and inspire many other studies in 3D vision. For example, the multi-modality of our method (Fig.\ref{fig:pair1}) can inspire research on image-based geometry editing; the practice of integrating geometry template into implicit functions can be used to improve the robustness of other 3D recovery tasks. Moreover, the semantic implicit functions can also be adopted in research on 3D perception, parsing and understanding. 


\begin{figure}
	\centering
	\includegraphics[width=\linewidth]{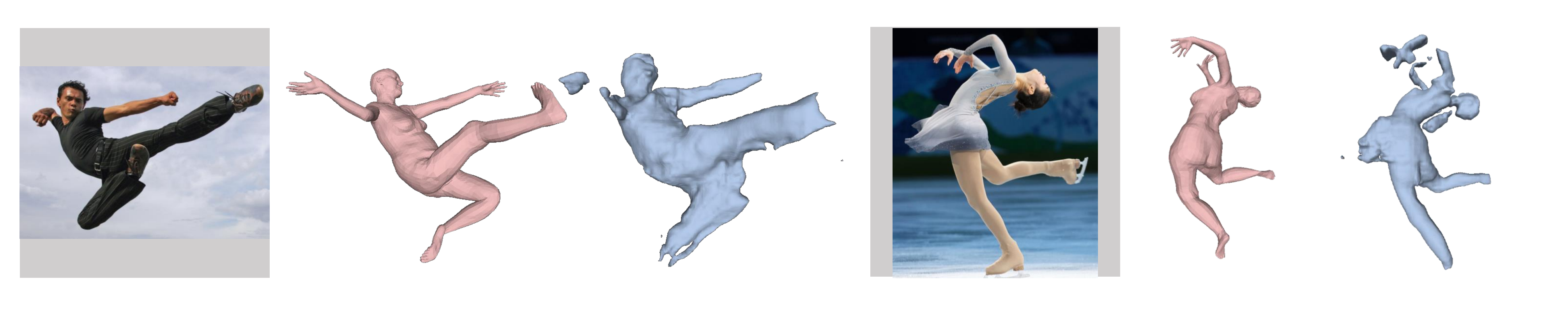}
	\caption{Failure cases. For extremely challenging poses, our method fails to generate plausible reconstruction results. }
	\label{fig:failure}
\end{figure}

\noindent\textbf{Limitation and Future Work. } 
Our method needs high-quality human scans for training. However, it is highly costly and time consuming to obtain a large-scale dataset of high-quality human scans. Moreover, the currently available scanners for human bodies require the subject to keep static poses in a sophisticated capturing environment, which makes them incapable to capture real-world human motions in the wild and consequently our training data is biased towards simple static poses like standing. Therefore, although the proposed method already makes a step forward in terms of generalization capability, it still fails in the cases of extremely challenging poses, see Fig.\ref{fig:failure}. One important future direction is to alleviate the reliance on ground-truth by exploring large-scale image and video dataset for unsupervised training.



\ifCLASSOPTIONcaptionsoff
  \newpage
\fi



%

\bibliographystyle{IEEEtran}
\bibliography{IEEEabrv,egbib}




%




\end{document}